\documentclass[letterpaper]{article} % DO NOT CHANGE THIS
\usepackage{aaai23}  % DO NOT CHANGE THIS
\usepackage{times}  % DO NOT CHANGE THIS
\usepackage{helvet}  % DO NOT CHANGE THIS
\usepackage{courier}  % DO NOT CHANGE THIS
\usepackage[hyphens]{url}  % DO NOT CHANGE THIS
\usepackage{graphicx} % DO NOT CHANGE THIS
\usepackage{bbm}
\usepackage{natbib}  % DO NOT CHANGE THIS AND DO NOT ADD ANY OPTIONS TO IT
\usepackage{amsfonts}
\usepackage{caption} % DO NOT CHANGE THIS AND DO NOT ADD ANY OPTIONS TO IT
\usepackage[ruled,vlined]{algorithm2e}
\usepackage{newfloat}
\usepackage{hyperref}

\usepackage{listings}
\DeclareCaptionStyle{ruled}{labelfont=normalfont,labelsep=colon,strut=off} % DO NOT CHANGE THIS
\usepackage{smile} 
\usepackage{xcolor,colortbl}
\usepackage{subfig}

\newif\ifsubmit
\submitfalse
\definecolor{darkpink}{rgb}{0.91, 0.33, 0.5}

\ifsubmit
\newcommand{\zc}[1]{}
\newcommand{\yue}[1]{}
\newcommand{\ran}[1]{}
\newcommand{\carl}[1]{}
\newcommand{\hejie}[1]{}
\newcommand{\xkan}[1]{}
\newcommand{\joyce}[1]{}
\newcommand{\yanqiao}[1]{}

\else
\newcommand{\zc}[1]{{\color{red}[Chao: #1]}}
\newcommand{\ran}[1]{{\color{teal}[Ran: #1]}}
\newcommand{\carl}[1]{{\color{cyan}[Carl: #1]}}  
\newcommand{\hejie}[1]{{\color{brown}#1}}
\newcommand{\joyce}[1]{{\color{darkpink}[Joyce: #1]}}
\newcommand{\xkan}[1]{{\color{orange}[Xuan: #1]}}
\newcommand{\yue}[1]{{\color{blue}[Yue: #1]}}
\newcommand{\yanqiao}[1]{\color{purple}[Yanqiao: #1]}
\fi

\usepackage{xspace}
\newcommand{\ours}{\sf {NeST}\xspace}

\title{Neighborhood-Regularized Self-Training for Learning with Few Labels}

\author {
Ran Xu\textsuperscript{\rm 1}, Yue Yu\textsuperscript{\rm 2}, Hejie Cui\textsuperscript{\rm 1}, Xuan Kan\textsuperscript{\rm 1}, Yanqiao Zhu\textsuperscript{\rm 3}, Joyce Ho\textsuperscript{\rm 1}, Chao Zhang\textsuperscript{\rm 2}, Carl Yang\textsuperscript{\rm 1}\thanks{Corresponding author.}
}
\affiliations {
    \textsuperscript{\rm 1} Emory University,
    \textsuperscript{\rm 2} Georgia Institute of Technology,
    \textsuperscript{\rm 3} University of California, Los Angeles\\
    \{ran.xu,hejie.cui,xuan.kan,joyce.c.ho,j.carlyang\}@emory.edu, \{yueyu,chaozhang\}@gatech.edu, yzhu@cs.ucla.edu \\
}

\begin{document}

\maketitle

\begin{abstract}
Training deep neural networks (DNNs) with limited supervision has been a popular research topic as it can significantly alleviate the annotation burden.
Self-training has been successfully applied in semi-supervised learning tasks, but one drawback of self-training is that it is vulnerable to the label noise from incorrect pseudo labels. 
Inspired by the fact that samples with similar labels tend to share similar representations, we develop a neighborhood-based sample selection approach to tackle the issue of noisy pseudo labels. 
We further stabilize self-training via aggregating the predictions from different rounds during sample selection. 
Experiments on eight tasks show that our proposed method outperforms the strongest self-training baseline with 1.83\% and 2.51\% performance gain for text and graph datasets on average. Our further analysis demonstrates that our proposed data selection strategy reduces the noise of pseudo labels by 36.8\% and saves 57.3\% of the time when compared with the best baseline. Our code and appendices will be uploaded to \url{https://github.com/ritaranx/NeST}.
\end{abstract}

\section{Introduction}
In the era of deep learning, neural network models have achieved promising performance in most supervised learning settings, especially when combined with self-supervised learning techniques~\cite{chen2020simple,devlin2018bert,Hu2020Strategies,zhu2022structure}. 
However, they still require a sufficient amount of labels to achieve satisfactory performances on many downstream tasks.   
For example, in the text domain, curating NLP datasets often require domain experts to read thousands of documents and carefully label them with domain knowledge. 
Similarly, in the graph domain, molecules are examples naturally represented as graphs, and characterizing their properties relies on  density functional theory (DFT)~\cite{cohen2012challenges} which often takes several hours. 
Such a dependency on labeled data is one of the barriers to deploy deep neural networks (DNNs) in real-world applications.

To better adapt the DNNs to target tasks with limited labels, one of the most popular approaches is \emph{semi-supervised learning} (SSL), which jointly leverages unlabeled data and labeled data to improve the model's generalization power on the target task~\cite{yang2021survey,wang2022usb}. 
Although generative models~\cite{gururangan2019variational} and consistency-based regularization~\cite{tarvainen2017mean,miyato2018virtual,xie2020uda} methods have been proposed for semi-supervised learning, they either suffer from the issue of limited representation power~\cite{tsai2022contrast} or require additional resources to generate high-quality augmented samples (e.g., for text classification, \citet{xie2020uda}~generate augmented
text via back-translation, which rely on a Machine Translation model trained with massive labeled sentence pairs).
% learn the implicit features of data to better model data distributions
Consequently, they cannot be readily applied to low-resource scenarios. 

Self-training is a proper tool to deal with the deficiency of labeled data via gradually enlarging the training set with pseudo-labeled data~\cite{rosenberg2005semi}.  
Specifically, it can be interpreted as a \emph{teacher-student framework}: 
the teacher model generates pseudo labels for
the unlabeled data, and the student model updates its parameters by minimizing the discrepancy between its predictions and the pseudo labels~\cite{xie2020self,mukherjee2020uncertainty}.
Though conceptually simple, self-training has achieved superior performance for various tasks with limited labels, such as image classification~\cite{sohn2020fixmatch,rizve2021in}, natural language understanding~\cite{du2020self}, sequence labeling~\cite{liang2020bond}, and graph learning~\cite{hao2020asgn}.
Self-training has also been successfully extended to other settings including weak supervision~\cite{zhang2021wrench} and zero-shot learning~\cite{li2022masked}.

However, one major challenge of self-training is that it suffers from \emph{confirmation bias}~\cite{arazo2020pseudo} --- when the teacher model memorizes some biases and generates incorrect pseudo labels, the student model will be reinforced to train with these wrong biases. As a result, the biases may amplify over iterations and deteriorates the final performance. 
To suppress the noisy pseudo labels in self-training, \citet{xu2021dp,zhang2021flexmatch,sohn2020fixmatch,kim2022conmatch} leverage model predictive confidence with a thresholding function, ~\citet{mukherjee2020uncertainty,tsai2022contrast} propose to leverage model uncertainty to select samples with low uncertainty, and \citet{wang2021meta} use meta-learning to conduct instance reweighting for sequence labeling. 
Although these approaches attempt to reduce the label noise, they select the data for self-training based on the model prediction only.  
However, the predictions of the deep neural network can be over-confident and biased~\cite{guo2017calibration,kong2020calibrated,kan2021zero}, and directly using such predictions without any intervention to filter pseudo labels cannot effectively resolve the label noise issue.  
Another problem from self-training is \emph{training instability}, 
as it selects pseudo-labeled data only based on the prediction of the \emph{current} round.
Due to the stochasticity involved in training neural networks (e.g., random initialization, training order), the prediction can be less stable~\cite{yu2022cold}, especially for the noisy pseudo-labeled data~\cite{xia2022sample}. 
Consequently, the noise in the previous rounds may propagate to later rounds, which deteriorate the final performance.

Motivated by the above, we propose {\ours}, a simple yet powerful approach guided by the data representations, to boost the performance of self-training for few-shot learning.
Inspired by recent works indicating that the representations from deep neural networks can be discriminative and less affected by noisy labels~\cite{li2021how}, 
we harness the features learned from the neural models to select the most reliable samples in self-training.
In addition, several works have indicated that samples within the same category tend to share similar representations, such as category-guided text mining~\cite{meng2020discriminative} and motif-driven graph learning~\cite{zhang2020motif}. 
Similarly, we hypothesize that a sample's pseudo label is more likely to be correct only if its prediction is similar to the neighbor labeled instances in the embedding space. 
To fulfill the denoising purpose, {\ours} creates the neighborhood for each unlabeled data by finding the top-$k$ nearest labeled samples, 
then calculates the divergence between its current prediction and the label of its neighbors to rank the unlabeled data. 
As a result, only the instances with the lowest divergence will be selected for self-training, which mitigates the issue of label noise.  
Moreover, to robustly select the training samples for self-training, we aggregate the predictions on different iterations to promote samples that have lower uncertainty over multiple rounds for self-training. 

We remark that {\ours} is an efficient substitution for existing self-training approaches and can be combined with various neural architectures. 
The contributions of this paper are:
\begin{itemize}
    \item We propose {\ours} to improve the robustness of self-training for learning with few labels only. 
    \item We design two additional techniques, namely neighborhood-regularized sample selection to reduce label noise, and prediction aggregation to alleviate the training instability issue.
    \item Experiments on 4 text datasets and 4 graph datasets with different volumes of labeled data verify that {\ours} improves the performance by 1.83\% and 2.51\% respectively and saves the running time by 57.3\%. 
\end{itemize}

\section{Related Work}
\label{sec:related}
Self-training is one of the earliest approaches to semi-supervised learning~\cite{rosenberg2005semi}.
The method uses a teacher model to generate new labels on which a student model is fitted.  
The major drawback of self-training is that it is vulnerable to label noise~\cite{arazo2020pseudo}. 
There are several popular approaches to stabilize the self-training process, such as using sample selection~\cite{mukherjee2020uncertainty,sohn2020fixmatch} and reweighting strategies~\cite{zhou2012self,wang2021meta} to filter noisy labels or designing noise-aware loss functions~\cite{liang2020bond,yu2022actune,tsai2022contrast} to improve the model's robustness against incorrectly labeled data. 
In addition, data augmentation methods~\cite{kim2022alp,chen2020mixtext,zhang2020seqmix} are also combined with self-training to improve the model's generalization ability.

Leveraging representation information has also been explored in semi-supervised learning. For example, \citet{li2021comatch,zhao2022lassl,yu2021fine} improve the representation via contrastive learning to assist semi-supervised learning.
Moreover, ACPL~\cite{liu2022acpl} and SimMatch~\cite{zheng2022simmatch} aggregates the labels from their neighbors in the feature space. 
While these approaches also attempt to harness sample representations, they do not directly denoise the pseudo labeled data for boosting the performance of self-training, which is the focus of our work. 
One concurrent work \cite{lang2022training} combines data representations with the cut statistic to select high quality training data. 
In particular, it aims to select reliable subsets directly from the weakly-labeled data. Instead, our work focuses on using clean labeled data to better denoise instances in self-training.

% \vspace{-0.3ex}

\section{Preliminaries}
\label{sec:prelim}
In this section, we first present the setup of semi-supervised learning and self-training, and then point out issues of the existing sample selection algorithms for self-training.

\subsection{Task Definition} 
In this paper, we study the semi-supervised learning problem, which is defined as follows. 
Given a few labeled $\mathcal{X}_l = \{(x_i,y_i)\}_{i=1}^L$ and unlabeled data $\mathcal{X}_u = \{x_j\}_{j=1}^U$ $(L \ll U)$, we seek to learn a predictor $f(x;\theta): \mathcal{X} \rightarrow \mathcal{Y}$. Here $\mathcal{X}=\mathcal{X}_l \cup \mathcal{X}_u$ denotes all the input data and $\mathcal{Y}$ is the label set, which can either be discrete (for classification) or continuous (for regression).
$f(x;\theta)$ is either a $C$-dimensional \emph{probability simplex} for classification where $C$ is the number of classes or a \emph{continuous value} for regression.

\subsection{Introduction to Self-training} 
Self-training can be interpreted as a teacher-student framework~\cite{mukherjee2020uncertainty,xie2020self}, with $\theta_t$ and $\theta_s$ denoting the teacher and student model, respectively. The  process of the self-training is in Alg.~\ref{alg:main}. We discuss the key components in self-training as belows.
\begin{algorithm}[t]
	\KwIn{Labeled and unlabeled samples $\mathcal{X}_l$, $\mathcal{X}_u$; Neural prediction model $f(\cdot; \theta)$; Unlabeled set $\hat{\cX_u}$; Number of self-training iterations $T$; Number of steps in each iteration $T_1$.}
	// \textit{Train the model on labeled data $\cX_l$ as initialization.} \\
	Update $\theta_s, \theta_t$ by Eq.~\ref{eq:init} using Adam. \\
	// \textit{Self-training.} \\
	\For{$t = 1, 2, \cdots, T$}{
	    Select $\hat{\cX_u^{t}} \ (|\hat{\cX_u^{t}}|=b)$ with $\theta_t$ by Eq.~\ref{eq:select}. \\
	    Adding $\hat{\cX_u^{t}}$ for self-training $\hat{\cX_u}$ = $\hat{\cX_u} \cup \hat{\cX_u^{t}}$. \\
		Update pseudo labels $\tilde{y}$ by Eq.~\ref{eq:soft} or~\ref{eq:regress}  for $\hat{\cX_u^t}$. \\
		\For{$k = 1, 2, \cdots, T_1$}{
			Sample a minibatch $\cB$ from $\hat{\cX_u}$. \\
            Update $\theta_s$ with loss $\cL$ in Eq.~\ref{eq:st} using Adam.
		}
		Update teacher model $\theta_t \leftarrow \theta_s$.
	}
	\KwOut{Final model $f(\cdot; \theta_s)$.}
	\caption{Procedures of Self-training. }
	\label{alg:main}
\end{algorithm}

\paragraph{Initialization of Models.} 
The labeled data $\mathcal{X}_l$ are used to initialize the models as $\theta_s^{(0)} = \theta_t^{(0)} = \theta_{\text{init}}$, where
\begin{align} \label{eq:init}
\setlength{\abovedisplayskip}{0.1pt}
\setlength{\belowdisplayskip}{0.1pt}
    \theta_{\text{init}} = \min_\theta ~\cL\textsubscript{sup}(\theta) =  \mathbb{E}_{(x_i, y_i) \in \cX_l} \ell\textsubscript{sup} \left( f(x_i; \theta), y_i \right).
\end{align}
$\ell\textsubscript{sup}(\cdot; \cdot)$ represents the supervised loss, which is the cross-entropy loss for classification and the mean squared error loss for regression.

\paragraph{Pseudo Label Generation with Teacher Model $\theta_t$.}
We use the teacher model's prediction $f(x;\theta_t)$ to generate pseudo labels for $\cX_u$.
For classification problems, the pseudo labels can be written as 
\begin{equation}
\tilde{\by}_{\text{hard},j}= \begin{cases}1, & \text { if } j = \underset{k \in \mathcal{Y}}{\text{argmax}}\left[{f}(x; \theta_t)\right]_k; \\ 0, & \text { else.} \end{cases}  
 \label{eq:soft}
\end{equation}

For the regression task, since the output is a continuous value, the teacher model's output is directly used as the pseudo label
\begin{equation}
\tilde{y}={f}(x; \theta_t).
\label{eq:regress}
\end{equation}

\paragraph{Sample selection.}
Directly using all the pseudo-labeled data for self-training often yields sub-optimal results, as the erroneous pseudo labels hurt the model performance. 
To mitigate this issue, recent works attempt to select only a subset of the unlabeled data for self-training. We denote the sample policies as $\psi(\cdot)$, which can be generally written as 
\begin{equation}
\hat{\cX_u}=\psi\left(\cX_u, f(x;\theta_t)\right).
\label{eq:select}
\end{equation}
We omit the superscript for simplicity. The common choice for $\psi(\cdot)$ including using predictive confidence~\cite{sohn2020fixmatch,zhang2021flexmatch} or model uncertainty~\cite{mukherjee2020uncertainty,tsai2022contrast}.

\paragraph{Model Training and Update.}
With the generated pseudo labels, we then train a student model ${\theta}_s$ to minimize the loss for both labeled and unlabeled data by solving
\begin{equation}
\min_{\theta_s} ~\lambda\cL\textsubscript{sup}({\theta_s}) + (1-\lambda) \mathbb{E}_{x_j \in \hat{\cX_u}} \ell\textsubscript{st} \left( f(x_j; {\theta_s}), \tilde{y}_j \right),
\label{eq:st}
\end{equation}
where  $\cL\textsubscript{sup}$ is defined in Eq.~\ref{eq:init}, $\hat{\cX_u}$ is obtained via Eq.~\ref{eq:select}, and $\ell\textsubscript{st}=\mathbbm{1}\{\left[f({x}_j;  \theta_s)\right]_{\tilde{y}_j}>\gamma \} \cdot \ell_{\text{sup}}$ is the loss function for unlabeled data with the thresholding function~\cite{sohn2020fixmatch,xie2020uda}. 
We iterate the process by treating the trained student model as the teacher to generate new pseudo labels and train a new student model based on the new generated labels until the model converges.

\begin{figure}
	\centering
	\subfloat{
		\includegraphics[width=0.32\linewidth]{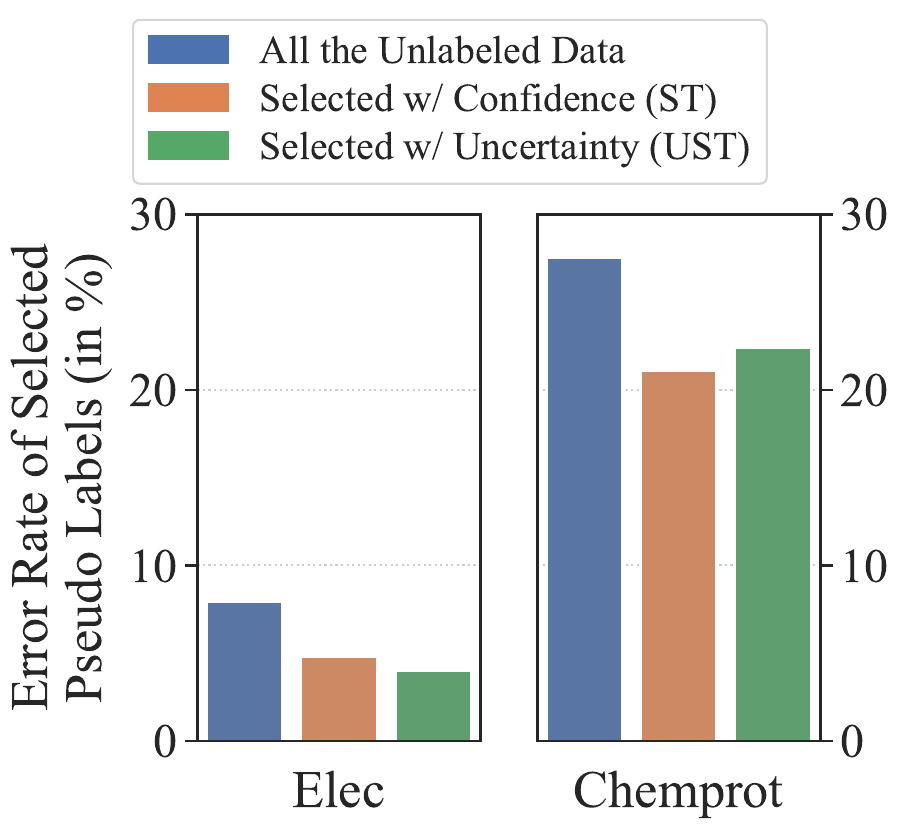}
		\label{fig:example_acc_pl}
	} %\hfill
	\subfloat{
		\includegraphics[width=0.32\linewidth]{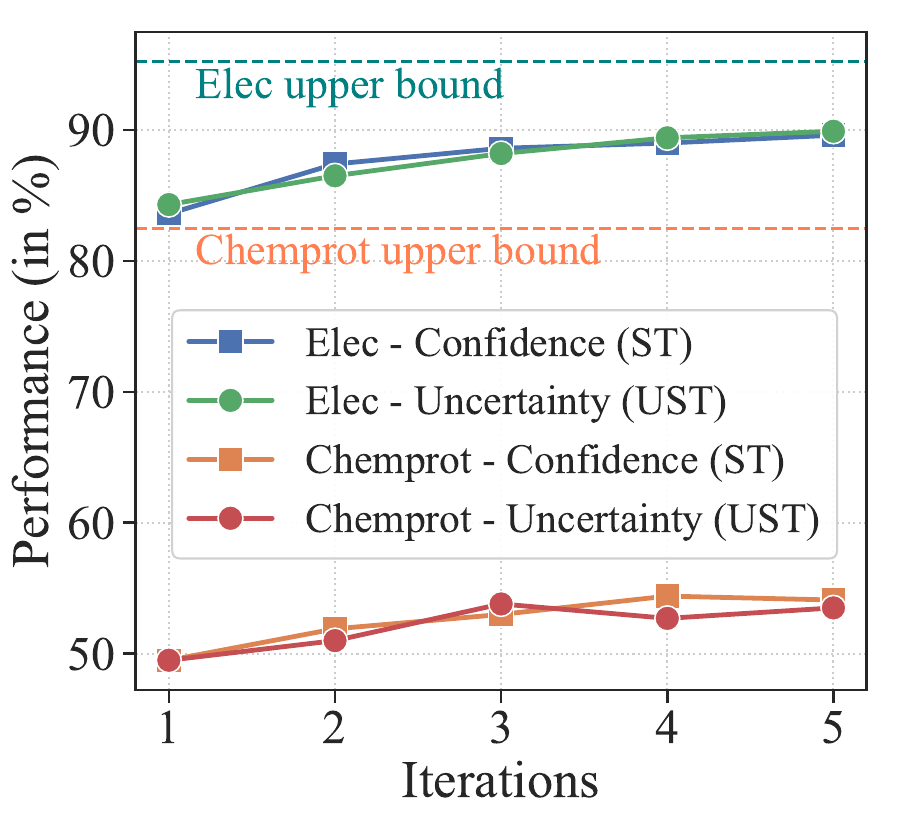}
		\label{fig:example_st}
	}
	\subfloat{
		\includegraphics[width=0.32\linewidth]{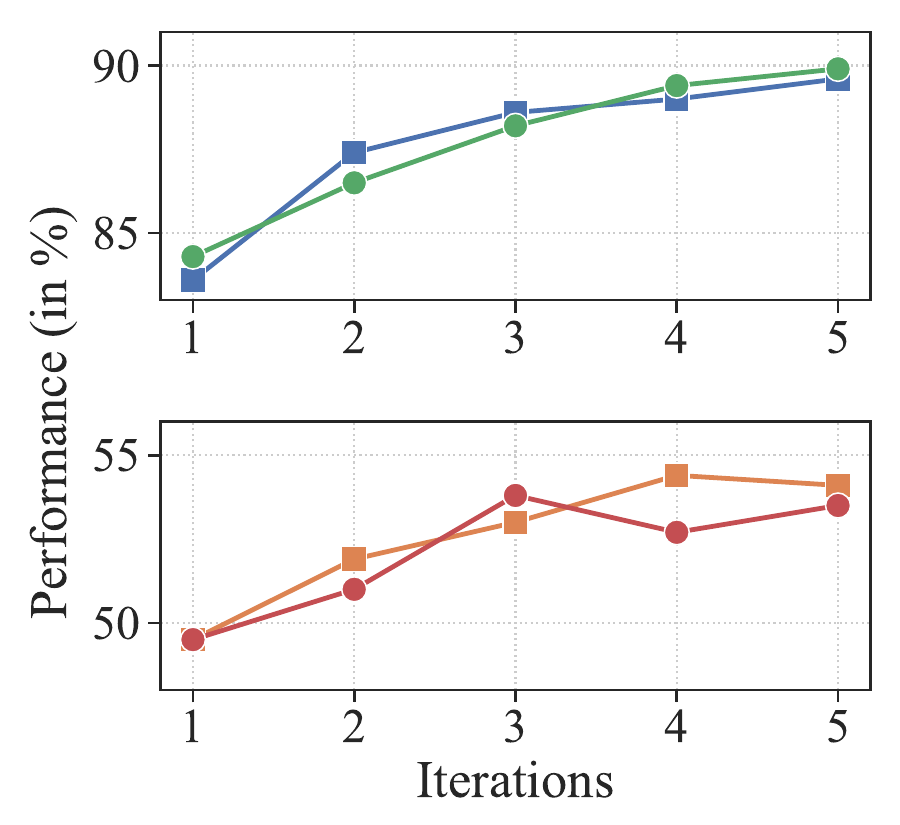}
		\label{fig:example_st}
	}
	\caption{\textbf{Left:} The average error rate of all pseudo labels and the selected pseudo labels with different selection strategies. \textbf{Middle:} The performance on different self-training iterations. The upper bounds are the accuracy with full clean labels. \textbf{Right:} Zoom in of the performance in the middle.}
	\vspace{-1ex}
\label{fig:analysis}
\end{figure}

\subsection{Challenges of Self-training}
To illustrate that the existing sample selection approaches are flawed and cannot resolve the label noise issue, we first demonstrate the performance of two widely-used selection criteria: \emph{predictive confidence} (ST~\cite{rosenberg2005semi,du2020self,liang2020bond}) and \emph{model uncertainty} (UST~\cite{mukherjee2020uncertainty}) for self-training. The details of these two approaches are discussed in Appendix~\ref{apd:select}. 
Note that, for these two approaches, we follow the original implementation to select the unlabeled set $\cX_u$ 
in each iteration. 
We use a binary sentiment classification dataset \emph{Elec} and a chemical relation extraction dataset \emph{Chemprot} with ten classes as an example for \emph{easier} and \emph{harder} task, respectively. 
For both datasets, we first train the model with 30 clean labels per class. 

Figure~\ref{fig:analysis} shows the error rate of pseudo labels selected following these two criteria. 
We observe that these two methods are effective on easier tasks, where the selected data has a relatively low error rate. It achieves comparable performance with the fully-supervised method (95\%) with less than 1\% of the clean labeled data. 
However, for more challenging tasks with a larger number of classes, the performance of the initial model may not be satisfactory. The error rate of pseudo labels increases up to 16\% on \emph{Chemprot} compared with \emph{Elec}. 
Consequently, the gap between semi-supervised learning and fully-supervised learning is even larger --- more than 25\% in terms of F1 score. 
This phenomenon suggests that the \emph{label noise issue} is still the major challenge that hampers the self-training performance. 
Moreover, using \emph{model uncertainty}~\cite{gal2016dropout} for sample selection does not fully address this challenge; the gain can be marginal on harder datasets.

Apart from the label noise, we also observe performance fluctuations over different self-training rounds. 
We name this as the \emph{training instability} issue, which occurs 
when the teacher model only looks at the previous round and memorizes the label noise specifically in that round. Then in the next iteration, the student model can easily overfit the noise.

% \vspace{-0.5ex}
\begin{figure}[t]
    \centering
    \includegraphics[width=0.98\linewidth]{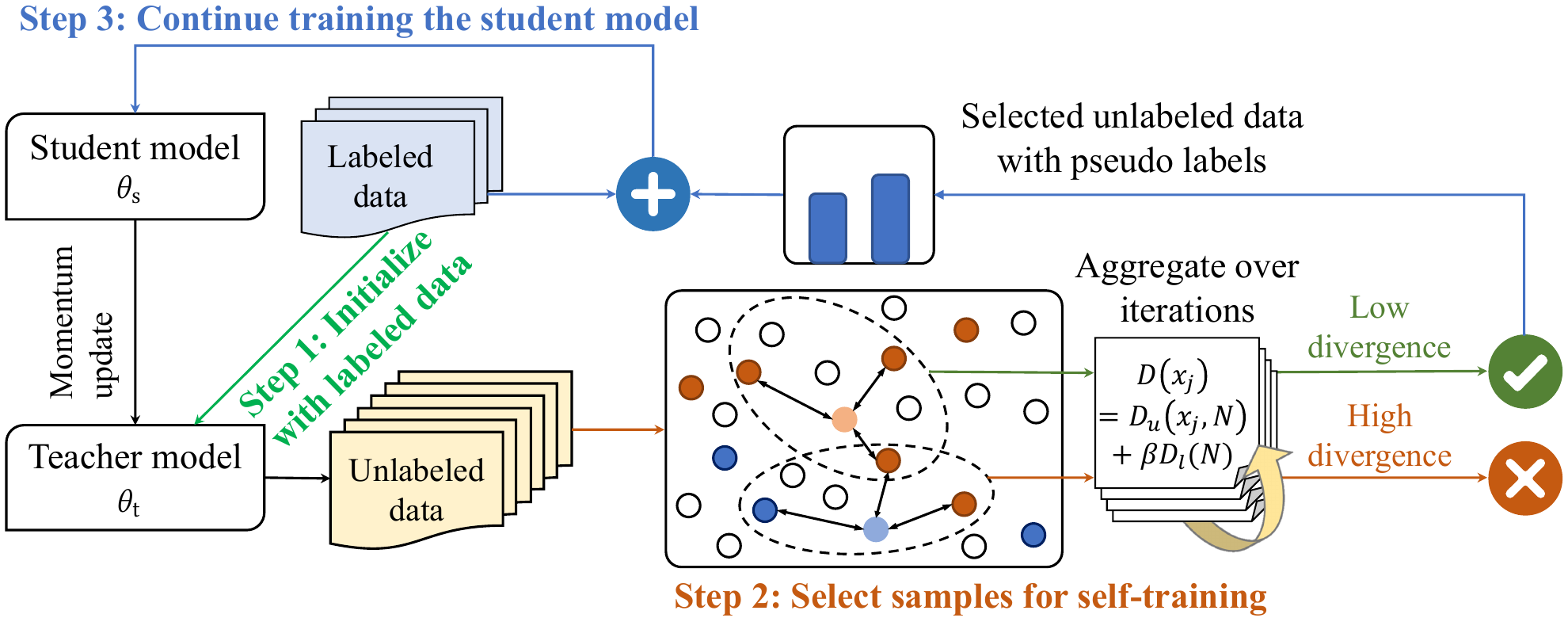}
    \caption{The framework of {\ours}. Red and blue points stand for labeled data with different labels. White points represent unlabeled data. Light red and blue points stand for predictions of unlabeled data.}
    \label{fig:model}
    \vspace{-2ex}
\end{figure}

\section{Method}
\label{sec:method}
We present {\ours} to improve the stability of self-training by tackling the challenges mentioned in the previous section. 
The overview of {\ours} is in Figure \ref{fig:model}. 
Notably, we focus on the \emph{sample selection} step (Eq.~\ref{eq:select}), and we propose two key components, namely neighborhood-regularized selection strategy and prediction aggregation to promote the performance. 
The details of the two designs will be discussed in Section~\ref{sec:sample} and Section~\ref{sec:selftrain} respectively.

\subsection{Neighborhood-regularized Sample Selection}
% \vspace{-0.5ex}
\label{sec:sample}

Prior works have demonstrated that leveraging embeddings from the deep neural networks can identify the noisy labeled data~\cite{zhu2022detecting}. 
Motivated by this, we propose to harness the similarities in the embedding space to mitigate the issue of erroneous pseudo labels in self-training. 

Concretely, for each unlabeled sample $\bx_j$ with representation $\bv_j$, we adopt the $k$-nearest neighbors (KNN) algorithm to find the most similar \emph{labeled} samples in the feature space:
\begin{equation}
\cN_j= \{x_i \mid x_i \in \cX_l \cap \text{KNN}(\bv_j, \cX_l, k) \},
\label{eq:knn}
\end{equation}
where $\text{KNN}(\bv_j, \cX_l, k)$ denotes $k$ {labeled} examples in $\cX_l$ that are nearest to $\bv_j$.

\paragraph{Divergence-based Sample Selection.} 
We then calculate the scores for unlabeled samples $x_j \in \cX_u$ based on the weighted divergence
\begin{equation}
\cD(x_j) = \cD_{\text{u}}(x_j, \cN)  + \beta \cD_{\text{l}}(\cN), 
\label{eq:diversity}
\end{equation}
where unlabeled divergence $\cD_{\text{u}}$ and labeled divergence $\cD_{\text{l}}$ are defined below, and $\beta$ is a hyperparameter. 
This score $\cD(x_j)$ will be further used for sample selection. 

\paragraph{Unlabeled Divergence $\cD_{\text{u}}$.}
For each sample $x_j$ in the unlabeled set with the neighbor set $\cN$, we calculate the divergence between the prediction of $x_j$ and labeled data in  $\cN$ as 
\begin{equation}
\cD_{\text{u}}(x_j, \cN) = \sum_{(x_i, y_i) \in \cN}d (f(x_j; \theta_t), y_i),
\label{eq:inter_diversity}
\end{equation}
where $d$ is the Kullback--Leibler (KL) divergence for classification and L2 distance for regression (same as follows). To interpret Eq.~\ref{eq:inter_diversity}, we note that samples having the prediction \emph{close} to the nearest labeled instances will have lower $\cD_{\text{u}}$. 

\paragraph{Labeled Divergence $\cD_{\text{l}}$.} It measures the divergence among the labels within the neighbor set $\cN$. We first calculate the average label $\overbar{y} = \sum_{(x_i, y_i) \in \cN}\frac{y_i}{|\cN|}$, and then measure the labeled divergence as
\begin{equation}
\cD_{\text{l}}(\cN) = \sum_{(x_i, y_i) \in \cN}d (\overbar{y}, y_i).
\label{eq:intra_diversity}
\end{equation}
For each group $\cN$, samples with similar labels will have smaller divergence $\cD_{\text{l}}(\cN)$.

To summarize, a low divergence score $\cD(x_j)$ indicates that the prediction of the unlabeled data point $x_j$ is \emph{close} to its neighbors, and the labels of its neighbors are \emph{consistent}. Thus, we hypothesize that such samples are more likely to be correct and use them for self-training.

\subsection{Robust Aggregation of Predictions from Different Iterations}
\label{sec:selftrain}
The results in Figure~\ref{fig:example_st} clearly demonstrate that only using the prediction on the current iteration for sample selection cause training instabilities.  
To effectively mitigate the bias in the current iteration and stabilize the self-training, we propose to exploit the training prediction at different training iterations more robustly~\cite{xia2022sample}. 
To achieve this, we aggregate the value $\cD^{(t)}(x_j)$ in the $t$-th round as 
\begin{equation}
    \mu^{(t)}(x_j) = (1-m) \times\mu^{(t-1)}(x_j)  + m \times \left(\cD^{(t)}(x_j)\right), 
    \label{eq:agg}
\end{equation}
where $m$ is a hyperparameter bounded between 0 and 1 that controls the weight for previous rounds. 

To interpret Eq.~\ref{eq:agg}, we argue that $\mu(x_j)$ will be small only when the model outputs \emph{consistently low} scores for a sample $x_j$ in different iterations of self-training, as the model is more certain about these samples. 
On the contrary, if the model gives inconsistent predictions in different iterations, then the model is potentially uncertain about the prediction, thus adding its pseudo label in the next iteration may hurt the self-training process.
Motivated by this idea, we remove the sample with inconsistent predictions over different iterations to further suppress noisy labels.

To put the above two strategies together, our policy for sample selection in the $t$-th round $\psi(\cdot)$  is mainly based on the value of   $\mu^{(t)}(x_j)$  in Eq.~\ref{eq:agg}. 
Specifically, in $t$-th iteration, we sample instances $x_j \in \cX_u$ without replacement using the probability

\begin{small}
\begin{equation}
p(x_j) \propto \frac{W-\mu^{(t)}(x_j)}{ \sum_{x_u \in \cX_u}\left(W-\mu^{(t)}(x_u) \right)},
\end{equation}
\end{small}
% \cite{ren2018learning,liu2020early,menon2021longtail}
where $W=\max_{x\in\cX_u}(\mu^{(t)}(x))$ is the normalizing factor.

\paragraph{Remark.} Our method introduces little computation overhead. For each unlabeled data, the neighborhood regularized sampling requires one extra kNN operation, which can be efficiently supported via FAISS \cite{faiss}. The $\mu^{(t)}(x_j)$ from previous iterations can be cached on disk and merged when selecting the training data for the new iteration. 
Other than the sample selection method $\psi(\cdot)$, {\ours} keeps other  components intact and can be plugged-in with any noise-robust learning techniques~\cite{menon2021longtail} and neural architectures.

% \vspace{-1ex}
\begin{table*}[t]
    \centering
    \renewcommand\arraystretch{0.95}
\resizebox{0.8\linewidth}{!}{
\begin{tabular}{cccccc}
\toprule
    \bf Dataset & \bf Domain & \bf Task & \bf \# Train / Test &\bf  \#  Class & \bf Metric  \\\midrule
    Elec     & Reviews  &  Sentiment Analysis & 25K / 25K & 2 & Acc. \\
    AG News  & News     &  Topic Classification & 120K / 7.6K &  4 & Acc. \\
    NYT      & News     &  Topic Classification & 30K / 3.0K  &  9 & Acc. \\
    Chemprot & Chemical   & Relation Classification & 12K / 1.6K &  10 & F1 \\ \midrule
    BBBP & Physiology  & Classification & 1.6k / 204 & 2 & ROC-AUC \\
    BACE & Biophysics  & Classification & 1.2k / 151 & 2 & ROC-AUC \\
    Esol & Physical Chemistry & Regression & 902 / 112 & --- & RMSE \\
    Lipophilicity & Physical Chemistry  & Regression & 3.3k / 420 & --- & RMSE \\
    \bottomrule
\end{tabular}
% \vspace{-1ex}
}
 \caption{Statistics of text and graph datasets.}
\vspace{-1.5ex}
\label{tab:dataset}
\end{table*}

\section{Experiments}
\label{sec:exp}
% In this section, we conduct extensive experiments to answer the
% following three research questions:
% % \noindent $\diamond$ 
% \textbf{RQ1}: How does {\ours} perform as compared with state-of-the-art methods?
% % \noindent $\diamond$ 
% \textbf{RQ2}: How does {\ours} perform with different backbones? 
% % \noindent $\diamond$ 
% \textbf{RQ3}: How do the key designs in {\ours} affect performance?

% \begin{table}[t]
%     \centering
%     \renewcommand\arraystretch{0.95}
% \resizebox{\linewidth}{!}{
% \begin{tabular}{cccccc}
% \toprule
%     \bf Dataset & \bf Category & \bf Task & \bf \# Train / Test & \bf  \#  Class & \bf Metric  \\\midrule
%     BBBP & Physiology  & Classification & 1.6k / 204 & 2 & ROC-AUC \\
%     BACE & Biophysics  & Classification & 1.2k / 151 & 2 & ROC-AUC \\
%     ESol & Physical Chemistry & Regression & 902 / 112 & --- & RMSE \\
%     Lipophilicity & Physical Chemistry  & Regression & 3.3k / 420 & --- & RMSE \\
%     \bottomrule
% \end{tabular}
% % \vspace{-1ex}
% }
%  \caption{Dataset statistics for molecule datasets.}
% \vspace{-2ex}
% \label{tab:dataset}
% \end{table}

% \vspace{-0.5ex}
\begin{table*}[t]
\centering
\renewcommand\arraystretch{0.95}
  \resizebox{0.97\linewidth}{!}{%
\begin{tabular}{lcccccccccccc}
\toprule
\multirow{2.5}{*}{Method} & \multicolumn{3}{c}{AG News \footnotesize(Accuracy, $\uparrow$)}    & \multicolumn{3}{c}{Elec \footnotesize(Accuracy, $\uparrow$)}    & \multicolumn{3}{c}{NYT \footnotesize(Accuracy, $\uparrow$)}& \multicolumn{3}{c}{Chemprot \footnotesize(F1, $\uparrow$) }            \\ \cmidrule(lr){2-4} \cmidrule(lr){5-7} \cmidrule(lr){8-10}  \cmidrule(lr){11-13}
& 30 & 50 & 100           & 30 & 50 & 100        & 30 & 50 & 100     & 30 & 50 & 100             \\ \midrule
 BERT~\shortcite{liu2019roberta} & 80.6\scriptsize±1.4 & 83.1\scriptsize±1.6 & 86.0\scriptsize±1.1 & 85.0\scriptsize±1.9 & 87.2\scriptsize±1.0 & 90.2\scriptsize±1.2 & 79.4\scriptsize±1.6 & 83.0\scriptsize±1.1 & 85.7\scriptsize±0.5 & 49.1\scriptsize±2.3 & 51.2\scriptsize±1.7 & 54.9\scriptsize±1.4 \\ 
  \midrule
 Mean-Teacher~\shortcite{tarvainen2017mean}  &  81.8\scriptsize±1.2 & 83.9\scriptsize±1.4 & 86.9\scriptsize±1.1 & 87.6\scriptsize±0.9  & 88.5\scriptsize±1.0 & 91.7\scriptsize±0.7 & 80.2\scriptsize±1.1 & 83.5\scriptsize±1.3 & 86.1\scriptsize±1.1 & 50.0\scriptsize±0.7 & 54.1\scriptsize±0.8 & 56.8\scriptsize±0.4 \\ 

VAT~\shortcite{miyato2018virtual} & 82.1\scriptsize±1.2 & 85.0\scriptsize±0.8& 87.5\scriptsize±0.9  &  87.9\scriptsize±0.8 & 89.8\scriptsize±0.5 & 91.5\scriptsize±0.4  &  80.7\scriptsize±0.7 & 84.4\scriptsize±0.9 & 
86.5\scriptsize±0.6 & 50.7\scriptsize±0.7&53.8\scriptsize±0.4&57.0\scriptsize±0.5\\ 

UDA~\shortcite{xie2020uda} & 86.5\scriptsize±0.9 & 87.1\scriptsize±1.2 & 87.8\scriptsize±1.2 & 89.6\scriptsize±1.1 &91.2\scriptsize±0.6 & 92.3\scriptsize±1.0  &--- & --- & --- & --- & ---  & ---  \\ 

MixText$^\dagger$~\shortcite{chen2020mixtext} & \underline{87.0\scriptsize±1.2}  & \underline{87.7\scriptsize±0.9} & 88.2\scriptsize±1.0 & 91.0\scriptsize±0.9  & 91.8\scriptsize±0.4 & 92.4\scriptsize±0.5 & --- & --- & --- & --- & ---  & ---  \\ 
\midrule
ST~\shortcite{rosenberg2005semi,liang2020bond}  & 86.0\scriptsize±1.4 & 86.9\scriptsize±1.0 & 87.8\scriptsize±0.6 & 89.6\scriptsize±1.2  & 91.4\scriptsize±0.4 & 92.1\scriptsize±0.5  &  \underline{85.4\scriptsize±0.9} & \underline{86.9\scriptsize±0.5} & \underline{87.5\scriptsize±0.5} & \underline{54.1\scriptsize±1.1}  & 55.3\scriptsize±0.7 &59.3\scriptsize±0.5 \\ 

UST~\shortcite{mukherjee2020uncertainty} & 86.9$^*$ & 87.4$^*$ & 87.9$^*$ & 90.0$^*$ & 91.6$^*$ & 91.9$^*$ &  85.0\scriptsize±0.6 & 86.7\scriptsize±0.4 & 87.1\scriptsize±0.3 & 53.5\scriptsize±1.3 & \underline{55.7\scriptsize±0.4} &\underline{59.5\scriptsize±0.7} \\ 

% MetaST~\shortcite{wang2021meta} & 86.6\scriptsize±0.5 & 87.5\scriptsize±0.9 &    \\  

CEST$^\ddagger$~\shortcite{tsai2022contrast} & 86.5$^*$ & 87.0$^*$ & \underline{88.4$^*$} & \underline{91.5$^*$} & \underline{92.1$^*$} & \underline{92.5$^*$} & --- & --- & --- & --- & --- & ---   \\  \specialrule{0.5pt}{0.5pt}{1pt} \midrule

{\ours}                                       & \textbf{87.8\scriptsize±0.8}     & \textbf{88.4\scriptsize±0.7}  & \textbf{89.5\scriptsize±0.3} & \bf 92.0\scriptsize±0.3 & \bf 92.4\scriptsize±0.2 & \bf 93.0\scriptsize±0.2 & \bf 86.5\scriptsize±0.7 & \bf 88.2\scriptsize±0.7  & \bf 88.6\scriptsize±0.6 &   \bf 56.5\scriptsize±0.7 & \bf 57.2\scriptsize±0.4  & \bf 62.0\scriptsize±0.5  \\ 
% \hline
% Gain $\Delta$  &  0.8 \scriptsize(0.9$\%$) &  0.7 \scriptsize(0.8$\%$) &  1.1 \scriptsize(1.2$\%$) &  0.4 \scriptsize(0.4$\%$) &  0.6 \scriptsize(0.7$\%$) &   0.4 \scriptsize(0.4$\%$) &  1.4 \scriptsize(1.6$\%$) &  1.5 \scriptsize(1.7$\%$) &  1.4 \scriptsize(1.6$\%$) &  2.4 \scriptsize(4.4$\%$) &  1.6 \scriptsize(2.9$\%$) &  3.1 \scriptsize(5.3$\%$)   \\  
\midrule
 Supervised & \multicolumn{3}{c}{93.0$^*$}& \multicolumn{3}{c}{95.3$^*$} & \multicolumn{3}{c}{93.6\scriptsize±0.5} & \multicolumn{3}{c}{82.5\scriptsize±0.4} \\
 \bottomrule
\end{tabular}
}
% \vspace{-1ex}
\caption{Performance on four datasets with various amounts of labeled data. The higher value always indicates better performance. 
\textbf{Bold} and \underline{underline} indicate the best and second best results for each dataset, respectively (Same as below). $^*$: The number is reported from the original paper. The implementation of CEST is \emph{not publicly available}. $\dagger$: The result is lower than the reported result in the original paper since they use a much larger development set. $\ddagger$: We remove the noise-aware loss as well as graph-based regularization for a fair comparison. The effect of these two terms is presented in table~\ref{tb:reg}.
}
\vspace{-1.ex}
\label{tb:main_text}
\end{table*}

% \vspace{-0.5ex}
\begin{table}[t]
\centering
\renewcommand\arraystretch{0.9}
  \resizebox{0.97\linewidth}{!}{%
\begin{tabular}{lcccccc}
\toprule
\multirow{2.5}{*}{Method} & \multicolumn{3}{c}{AG News \footnotesize(Accuracy, $\uparrow$)}    & \multicolumn{3}{c}{Elec \footnotesize(Accuracy, $\uparrow$)}          \\ \cmidrule(lr){2-4} \cmidrule(lr){5-7} 
& 30 & 50 & 100           & 30 & 50 & 100          \\ 
\midrule
CEST  & 86.5 & 87.5 & 88.4 & 91.5 & 92.1 & 92.5    \\ 
CEST w/ NRL & 87.1& 88.0 & 88.9 & 92.2  & 92.4 & 92.8  \\
% \specialrule{0.5pt}{0.5pt}{1pt} 
\midrule
{\ours}                                       &  {87.8\scriptsize±0.8}     & {88.4\scriptsize±0.7}  & {89.5\scriptsize±0.3}  & 92.0\scriptsize±0.3 & 92.4\scriptsize±0.2 & 93.0\scriptsize±0.2 \\ 
% \hline
{\ours} w/ NRL & \bf{88.3\scriptsize±0.5}   & \bf{88.9\scriptsize±0.6}  & \bf{89.8\scriptsize±0.2}  & \bf92.3\scriptsize±0.2 &\bf 92.7\scriptsize±0.2 &\bf 93.1\scriptsize±0.3
\\ \bottomrule
\end{tabular}
}
% \vspace{-1ex}
\caption{Performance comparison of CEST~\cite{tsai2022contrast} and {\ours} with noise-robust loss functions (NRL).
}
\vspace{-1.5ex}
\label{tb:reg}
\end{table}

% \vspace{-0.5ex}
\begin{table*}[t]
\centering
\renewcommand\arraystretch{0.95}
  \resizebox{0.97\linewidth}{!}{%
\begin{tabular}{lcccccccccccc}
\toprule
\multirow{2.5}{*}{Method} & \multicolumn{3}{c}{BBBP \footnotesize(ROC-AUC, $\uparrow$)}    & \multicolumn{3}{c}{BACE \footnotesize(ROC-AUC, $\uparrow$)}    & \multicolumn{3}{c}{Esol \footnotesize(RMSE, $\downarrow$)}& \multicolumn{3}{c}{Lipo \footnotesize(RMSE, $\downarrow$) }            \\ \cmidrule(lr){2-4} \cmidrule(lr){5-7} \cmidrule(lr){8-10}  \cmidrule(lr){11-13}
& 30 & 50 & 100           & 30 & 50 & 100        & 30 & 50 & 100     & 30 & 50 & 100             \\ \midrule
 Grover~\shortcite{rong2020grover} & 69.0\scriptsize±1.9 & 79.3\scriptsize±2.2 &  86.4\scriptsize±1.1  &  68.6\scriptsize±1.1  & 75.2\scriptsize±3.1  & 78.1\scriptsize±1.5  &  1.573\scriptsize±0.061  &  1.340\scriptsize±0.028  &  1.147\scriptsize±0.022  & 1.209\scriptsize±0.025  &  1.133\scriptsize±0.036  &  1.088\scriptsize±0.020\\ 
  \midrule
 Mean-Teacher~\shortcite{tarvainen2017mean}  &  71.8\scriptsize±1.2 & 80.9\scriptsize±1.4 & 87.2\scriptsize±0.6 & 69.5\scriptsize±0.8 & 76.6\scriptsize±0.7 & 80.2\scriptsize±0.2 & 1.500\scriptsize±0.018 & 1.298\scriptsize±0.020 & 1.104\scriptsize±0.021 & 1.154\scriptsize±0.027 & 1.091\scriptsize±0.018 & 1.067\scriptsize±0.035  \\ 

VAT~\shortcite{miyato2018virtual} & 72.2\scriptsize±1.0 & 80.0\scriptsize±1.6 & 88.5\scriptsize±0.5 & 69.8\scriptsize±0.5 & 76.5\scriptsize±0.4 & 78.6\scriptsize±0.8 & 1.492\scriptsize±0.039 & 1.264\scriptsize±0.031 & 1.056\scriptsize±0.014 & 1.199\scriptsize±0.016 & 1.089\scriptsize±0.021 & 1.050\scriptsize±0.027 \\ 

ASGN~\shortcite{hao2020asgn} & 72.1\scriptsize±1.3 & 80.5\scriptsize±1.3 & 87.3\scriptsize±0.9 & 69.6\scriptsize±1.0 & 77.2\scriptsize±1.0 & 79.3\scriptsize±0.4 & 1.449\scriptsize±0.030 & 1.249\scriptsize±0.018 & 1.096\scriptsize±0.004 & 1.123\scriptsize±0.033 & 1.084\scriptsize±0.039 & 1.023\scriptsize±0.032      \\ 

InfoGraph~\shortcite{sun2020infograph} &  \underline{72.5\scriptsize±0.9} & 81.2\scriptsize±0.3 & 88.5\scriptsize±0.5 & \underline{70.2\scriptsize±0.6} & 77.8\scriptsize±0.8 & \underline{80.4\scriptsize±0.5} & 1.414\scriptsize±0.041 & 1.222\scriptsize±0.017 & 1.082\scriptsize±0.013 & 1.125\scriptsize±0.024 & 1.082\scriptsize±0.027 & 1.039\scriptsize±0.020    \\ 
\midrule
ST~\shortcite{rosenberg2005semi}  & 71.0\scriptsize±2.0 & 80.4\scriptsize±1.4 & 87.8\scriptsize±1.4 & 67.9\scriptsize±1.3 & 75.8\scriptsize±2.0 & 78.9\scriptsize±1.0 & 1.463\scriptsize±0.043 & 1.225\scriptsize±0.030 & 1.105\scriptsize±0.025 & 1.135\scriptsize±0.047 & 1.090\scriptsize±0.043 & 1.030\scriptsize±0.013   \\ 

UST~\shortcite{mukherjee2020uncertainty} & 71.7\scriptsize±1.1 & \underline{81.8\scriptsize±0.7} & \underline{88.8\scriptsize±0.4} & 69.9\scriptsize±0.3 & \underline{78.0\scriptsize±0.4} & \underline{80.4\scriptsize±0.5} & \underline{1.408\scriptsize±0.026} & \underline{1.174\scriptsize±0.034} & \underline{1.023\scriptsize±0.010}& \underline{1.115\scriptsize±0.020} & \underline{1.069\scriptsize±0.014} & \underline{1.010\scriptsize±0.009} \\ 

% MetaST~\shortcite{wang2021meta} & 72.3\scriptsize±0.9 & 81.5\scriptsize±0.9 &  \blue{89.1\scriptsize±0.5} & 70.1\scriptsize±0.4 & \blue{78.2\scriptsize±0.3} &  \blue{80.7\scriptsize±0.2} & \blue{1.408\scriptsize±0.026} & \blue{1.186\scriptsize±0.042} & \blue{1.023\scriptsize±0.010} & \blue{1.115\scriptsize±0.020} & \blue{1.060\scriptsize±0.017} & \blue{1.010\scriptsize±0.010}   \\  
\specialrule{0.5pt}{0.5pt}{1pt}  \midrule
{\ours}                                       & \textbf{75.4\scriptsize±1.0}     & \textbf{83.5\scriptsize±0.8}  & \textbf{90.0\scriptsize±0.4} & \textbf{70.5\scriptsize±0.2}      & \textbf{79.3\scriptsize±0.3}  & \textbf{81.6\scriptsize±0.3} & \textbf{1.325\scriptsize±0.024}     & \textbf{1.130\scriptsize±0.026}   & \textbf{1.001\scriptsize±0.002} & \textbf{1.088\scriptsize±0.011} & \textbf{1.039\scriptsize±0.021}  & \textbf{0.992\scriptsize±0.013}  \\ 
% \hline
% Gain $\Delta$  &  2.9 \scriptsize(4.0$\%$) &  1.7 \scriptsize(2.1$\%$) &  1.2 \scriptsize(1.4$\%$) &  0.3 \scriptsize(0.6$\%$) &  1.3 \scriptsize(1.7$\%$) &   1.2 \scriptsize(1.5$\%$) &  0.083 \scriptsize(5.9$\%$) &  0.044 \scriptsize(3.7$\%$) &  0.022 \scriptsize(2.2$\%$) &  0.027 \scriptsize(2.4$\%$) &  0.030 \scriptsize(2.8$\%$) &  0.018 \scriptsize(1.8$\%$)   \\ 
\midrule
 Supervised$^\diamond$~\shortcite{rong2020grover} & \multicolumn{3}{c}{93.6}& \multicolumn{3}{c}{87.8} & \multicolumn{3}{c}{0.888} & \multicolumn{3}{c}{0.563} \\
 \bottomrule
\end{tabular}
}
% \vspace{-1ex}
\caption{Performance on four datasets with various amounts of labeled data. For classification datasets (BBBP, BACE), the higher value indicates better performance, while for regression datasets (Esol, Lipo), the lower value stands for better performance. 
\textbf{Bold} and \underline{underline} indicate the best and second best results, respectively (Same as below). 
}\label{tab:Main}
\vspace{-1ex}
\label{tb:main}
\end{table*}
\subsection{Experiment Setup}

We conduct experiments for semi-supervised learning on eight datasets
% two tasks, namely text classification and molecular property prediction (graph-level) 
to demonstrate the efficacy of {\ours}. 
Four of them are text-related tasks, including text classification and relation extraction. We employ the pre-trained BERT
% \texttt{BERT-base-uncased}~\cite{devlin2018bert} 
from the HuggingFace~\cite{wolf2019huggingface} codebase for the implementation. 
The other four are graph-based tasks, where we choose molecular property prediction as the main task and use pre-trained \texttt{Grover-base} 
\footnote{Since we do not focus on developing better self-supervised learning methods, we do not compare with other GNN pretraining models~\cite{you21a,zhu2021graph}}
\cite{rong2020grover} as the backbone. 
The same backbone is used for both {\ours} and baselines to ensure a fair comparison. 
% four widely used datasets for molecular property prediction

\paragraph{Semi-supervised Learning Settings.}  
For each dataset, we train our method and baselines with different numbers of labeled data from \{30, 50, 100\} per class. The remaining in the training set is considered as \emph{unlabeled data}. 
As suggested by~\citet{bragg2021flex}, we keep the size of the validation set to be the same as the number of labeled data to simulate the realistic setting.  
% \hejie{didn't get it}.\ran{better?}
For each dataset, we apply 3 runs on 3 splits and report the mean and standard deviations. 

\paragraph{Parameter Settings.}  
% In our main experiments, we use the pre-trained Grover-base\footnote{Since our focus is not on developing better self-supervised learning methods, we do not compare with other graph neural network pretraining approaches such as~\cite{you2020graph,you21a}.}\cite{rong2020grover}, the state-of-the-art model on MoleculeNet as the backbone for both {\ours} and baselines. 
We use Adam~\cite{kingma2014adam} as the optimizer and tune the learning rate in \{1e-5, 2e-5, 5e-5\}. The batch size is selected from \{8, 16, 32\}. 
Other hyperparameters in {\ours} include $T, T_1, \gamma$ for self-training,  $\beta, b, k$ for sample selection in Eq.~\ref{eq:diversity}, and $\lambda$ in Eq.~\ref{eq:st}.  %and $\tau$ for soft-labeling in Eq.~(\ref{eq:soft}). 
We set $\beta=0.1$, $\gamma=0.9$, $\lambda=0.5$, $m=0.6$, $T=5$, $T_1=1000$ for all datasets, and tune $b=c|\cX_l|$ with $c\in\{3, 5, 10, 20\}$ for text datasets and $c\in\{1, 3, 5\}$ for graph datasets. We study the effect of $k$ and $c$ in Section~\ref{sec:abla}. Details for each dataset are in Appendix~\ref{apd:param}.

\paragraph{Baselines.}  
We compare {\ours} with the following baselines. We use $\dagger$ to represent baselines designed for text-based tasks and $\ddagger$ to represent baselines for graph-based tasks.
\begin{itemize}
    \item \textbf{BERT}$^\dagger$~\cite{devlin2018bert,lee2020biobert} is the supervised baseline for text-based tasks.
    \item \textbf{Grover}$^\ddagger$~\cite{rong2020grover} is the supervised baseline for molecular property prediction.
    \item \textbf{{Mean-Teacher} (MT)}~\cite{tarvainen2017mean} updates the teacher model as a moving average of the student model's weight and adds a consistency regularization between the student and teacher model.
    \item \textbf{Virtual Adversarial Training (VAT)}~\cite{miyato2018virtual} adds a regularization term between the sample with the adversarial noise and its prediction.
    \item  \textbf{Self-training (ST)}~\cite{rosenberg2005semi} is a conventional self-training method that adds \emph{most confident} pseudo labeled data to labeled set.
    \item  \textbf{UST}~\cite{mukherjee2020uncertainty}  selects data with \emph{lowest uncertainty} using MC-dropout for self-training.
    \item \textbf{UDA}$^\dagger$~\cite{xie2020uda} adopts back translation and TF-IDF word replacement as the data augmentation and adds consistency loss on predictions on the augmented data.
    \item \textbf{MixText}$^\dagger$~\cite{chen2020mixtext} interpolates  training data in the hidden space via Mixup as the data augmentation to improve model performance. 
    \item \textbf{CEST}$^\dagger$~\cite{tsai2022contrast} improves the UST method by designing the contrastive loss over sample pairs and noise-aware loss function.
    \item \textbf{{InfoGraph}}$^\ddagger$~\cite{sun2020infograph} is a semi-supervised graph classification method via maximizing mutual information between graph and substructures.
     \item \textbf{{ASGN}}$^\ddagger$~\cite{hao2020asgn} is a semi-supervised molecular property prediction that jointly exploits information from molecular structure and overall distribution. 
\end{itemize}

\subsection{Semi-supervised Learning on Text}
\paragraph{Datasets.} We conduct experiments on four widely used datasets in NLP.
We adopt Elec~\cite{mcauley2013hidden} for sentiment classification, AGNews~\cite{zhang2015character} and NYT~\cite{meng2020discriminative} for topic classification,
and Chemprot~\cite{taboureau2010chemprot} for chemical relation extraction in this set of experiments.
The statistics and evaluation metrics for each dataset are shown in Table~\ref{tab:dataset}. 
We use \texttt{BioBERT}~\cite{lee2020biobert} as the backbone for Chemprot as it is a domain specific dataset~\cite{cui2022can} and use \texttt{RoBERTa-base} for other datasets.

\paragraph{Results.} Table~\ref{tb:main_text} summarizes the experimental results on text datasets. We observe that {\ours} outperforms all baselines across all the four datasets under different volumes of labeled data, and the performance gain compared to the best baseline is around 1.83\% on average. 
Note that UDA and MixText require \emph{additional} data augmentation, which can be computationally expensive. 
Instead, {\ours} does not leverage any external resources but achieves better performance.  

For other self-training baselines, we observe that they cannot outperform our proposed method. As we keep other components unchanged, the gain is mainly due to the pseudo-labeled data denosing benefit of {\ours}. We will illustrate this in Section~\ref{sec:case}.
We also notice that the performance gain is more prominent on NYT and Chemprot datasets which have more classes, indicating {\ours} can be better adapted to tasks with fine-grained classes.

\paragraph{Incorporating Noise-robust Loss Functions.} To demonstrate {\ours} can be combined with other loss functions, Table~\ref{tb:reg} further compares {\ours} with CEST~\cite{tsai2022contrast} which adds additional noise-aware loss~\cite{menon2021longtail} and graph-based regularization.
The results show that these components can further improve the performance of {\ours}. Under both settings, {\ours} outperforms CEST, which justifies the efficacy of our proposed strategies.

\subsection{Semi-supervised Learning on Graphs}
\paragraph{Datasets.} We choose molecular property prediction as the target task for graph classification.
We conduct experiments on four widely used datasets from the MoleculeNet~\cite{wu2018moleculenet}, including BBBP~\cite{bbbp}, BACE~\cite{bace}, Esol~\cite{delaney2004esol} and Lipophilicity~\cite{lipo}. {The statistics and evaluation metrics for each dataset are shown in Table~\ref{tab:dataset}}. 
% We adopt the scaffold split to determine train/validation/test. 

\paragraph{Experiment results.} 
From the results in Table~\ref{tab:Main}, we can see that {\ours} outperforms all the baselines on all the datasets. In particular, the performance gain compared to the best baseline is around 2.5\% on average. Compared to the  Grover model using labeled data only, the gain is around 8.5\% on average. Notice that the traditional self-training method (ST) sometimes performs even worse than Grover fine-tuned on labeled data only, because confidence-based selection introduces large label noise, which leads to many wrong predictions. With proper control of noisy pseudo labels, UST generally outperforms other baselines.  
% which indicates that for molecular property prediction tasks, self-training is more suitable compared with consistency-based SSL approaches (e.g. MT, VAT). 
However, since they do not consider neighbor information, their performance is not as good as {\ours}.

\paragraph{Adapting {\ours} to different backbones.}
\label{sec:gat}
We use two datasets as an example to demonstrate that {\ours} can be adapted to different backbones.
Table~\ref{tab:adapt} shows the results of training an AttentiveFP~\cite{xiong2019pushing}, another popular GNN backbone based on graph attention networks for molecular property prediction.
Unlike Grover, AttentiveFP is not pre-trained on massive molecules, but trained from scratch in our experiments. 
We can see that the performance of AttentiveFP is worse than Grover in most cases. 
A key reason is that pre-trained Grover has considerably more parameters than AttentiveFP, and incorporates rich domain knowledge with self-supervised learning on unlabeled molecules. 
Nevertheless, {\ours} still outperforms all the baselines by 1.5\% on two datasets. 
This indicates that {\ours} does not rely on any specific architecture, and it serves as an effective plug-in module for different GNN models.

% We can see that the model trained from scratch performs worse than fine-tuning Grover (Table~\ref{tb:main}) in most cases. This is because AttentiveFP has significantly less parameters than RoBERTa, and is not pre-trained on massive text corpora.  
% Nevertheless,  {\ours} consistently outperforms the baseline methods.
% This indicates that our method is architecture independent, and does not rely on transferring existing semantic information. 
% As such, {\ours} serves as an effective plug-in module for existing models. 
% \yue{add discuss on gain is smaller, {\ours} requires a better representation.}

\begin{table}[t]
\centering
\renewcommand\arraystretch{0.95}
  \resizebox{\linewidth}{!}{%
\begin{tabular}{lcccccc}
\toprule
\multirow{2.5}{*}{Method} & \multicolumn{3}{c}{BBBP \footnotesize(ROC-AUC, $\uparrow$)}    & \multicolumn{3}{c}{BACE \footnotesize(ROC-AUC, $\uparrow$)}           \\ \cmidrule(lr){2-4} \cmidrule(lr){5-7}
& 30 & 50 & 100           & 30 & 50 & 100               \\ \midrule
 AttentiveFP~\shortcite{xiong2019pushing} &  66.4\scriptsize±1.3 & 75.6\scriptsize±1.1 & 82.9\scriptsize±0.5 & 68.0\scriptsize±1.4 & 72.2\scriptsize±1.3 & 74.0\scriptsize±0.9   \\ 
  \midrule
 Mean-Teacher~\shortcite{tarvainen2017mean}  & 67.9\scriptsize±0.7 & 76.3\scriptsize±0.3 & 83.4\scriptsize±0.2 & 70.1\scriptsize±0.3 & 73.8\scriptsize±0.3 & 75.5\scriptsize±0.6    \\ 

VAT~\shortcite{miyato2018virtual}  & 68.3\scriptsize±0.4 & 76.7\scriptsize±0.5 & 84.0\scriptsize±0.5 & 70.5\scriptsize±0.3 & 73.4\scriptsize±0.5 & 76.0\scriptsize±0.4    \\ 

ASGN~\shortcite{hao2020asgn}  & 70.0\scriptsize±0.4 & 77.1\scriptsize±0.2 & 84.1\scriptsize±0.3 & \underline{70.9\scriptsize±0.5} & 74.9\scriptsize±0.7 & 77.9\scriptsize±0.6        \\ 

InfoGraph~\shortcite{sun2020infograph}  & 68.5\scriptsize±0.4 & 79.3\scriptsize±0.8 & 83.8\scriptsize±0.4 & 70.7\scriptsize±0.3 & 75.3\scriptsize±0.2 & \underline{78.8\scriptsize±0.3} \\   
\midrule
ST~\shortcite{rosenberg2005semi}  &  69.8\scriptsize±0.2 & 76.0\scriptsize±1.4 & 84.0\scriptsize±1.1 & 67.5\scriptsize±0.5 & 71.4\scriptsize±3.4 & 76.0\scriptsize±0.8 \\ 

UST~\shortcite{mukherjee2020uncertainty}  & \underline{70.7\scriptsize±0.3} & \underline{79.2\scriptsize±0.5} & \underline{84.5\scriptsize±0.7} & 70.3\scriptsize±0.4 & \underline{75.5\scriptsize±0.3} & 78.7\scriptsize±0.6 \\ 
% MetaST~\shortcite{wang2021meta} & 70.4\scriptsize±0.9 & 79.0\scriptsize±0.7 & 84.1\scriptsize±0.6 & 70.6\scriptsize±0.5 & 75.3\scriptsize±0.6 & \underline{79.2\scriptsize±0.5}\\
\specialrule{0.5pt}{0.5pt}{1pt} \midrule
{\ours} & \textbf{71.2\scriptsize±0.4} & \textbf{80.7\scriptsize±0.5}  & \textbf{85.2\scriptsize±0.5} & \textbf{72.2\scriptsize±0.4}      & \textbf{76.6\scriptsize±0.5}  & \textbf{80.7\scriptsize±0.5}    \\ 
% \hline
% Gain $\Delta$  &  0.5 \scriptsize(0.7\%) &  1.5 \scriptsize(1.9\%) &  0.7 \scriptsize(0.8\%) &  1.3 \scriptsize(1.8\%) &  1.1 \scriptsize(1.5\%) &   1.9 \scriptsize(2.4\%)\\
% Grover w. gold labels                              &    \\ 
\bottomrule
\end{tabular}
}
% \vspace{-0.1ex}
\caption{Performance on two datasets with various amounts of labeled data using AttentiveFP as the backbone.}\label{tab:adapt}
\vspace{-1.5ex}
\end{table}

% \vspace{-0.3ex}
% \subsection{Adapting {\ours} to Different Backbones (RQ2)}
% \vspace{-0.3ex}

% \vspace{-0.5ex}
\subsection{Parameter and Ablation Studies}
% \vspace{-0.2ex}
\label{sec:abla}
We study the effect of different parameters of {\ours} on NYT and Chemprot with $30$ labels per class, shown in Figure~\ref{fig:ablation}. Results on  other datasets are in Appendix~\ref{apd:abla}. The performance first improves as $k$ increases, because larger $k$ allows more labeled data in each neighborhood, introducing less randomness and regularizing divergence calculation. When $k > 7$, the performance drops as the neighborhood includes labeled data far away, no longer serving as an appropriate regularizer. 
Similarly, as $c$ gets larger, the performance first increases and then decreases. This indicates that when $c$ is too small, the data are insufficient to train an effective student model, and when $c$ is too large, unlabeled data tend to be noisier and can hurt the performance.
% For the effect of $m$, we find that the model performance peaks when $m$ is around 0.95. The reason is that the teacher model will update too aggressively with a smaller $m$, while too slowly when $m$ is too large. 
% \joyce{One question might be why do a non-linear update (Eq. 11) rather than a simple averaging of the previous iterates?}
% The teacher model updates too aggressively with a smaller α (e.g., α = 0.7), and too conservatively with a larger alpha (e.g., α = 0.99).

We also inspect components of {\ours}, shown in Figure~\ref{fig:abl}. It is observed that both two strategies help to improve performance. The aggregation module stabilizes the self-training as the fluctuation issue has been mitigated. 
% \yue{hard label vs soft label}
% \yue{Effect of Noise-aware loss function, teacher/student model for prediction}
% \yue{Effect of $k$, $c$, $m$}

\begin{figure}
	\centering
	\vspace{-1ex}
	\subfloat[$k$]{
		\includegraphics[width=0.315\linewidth]{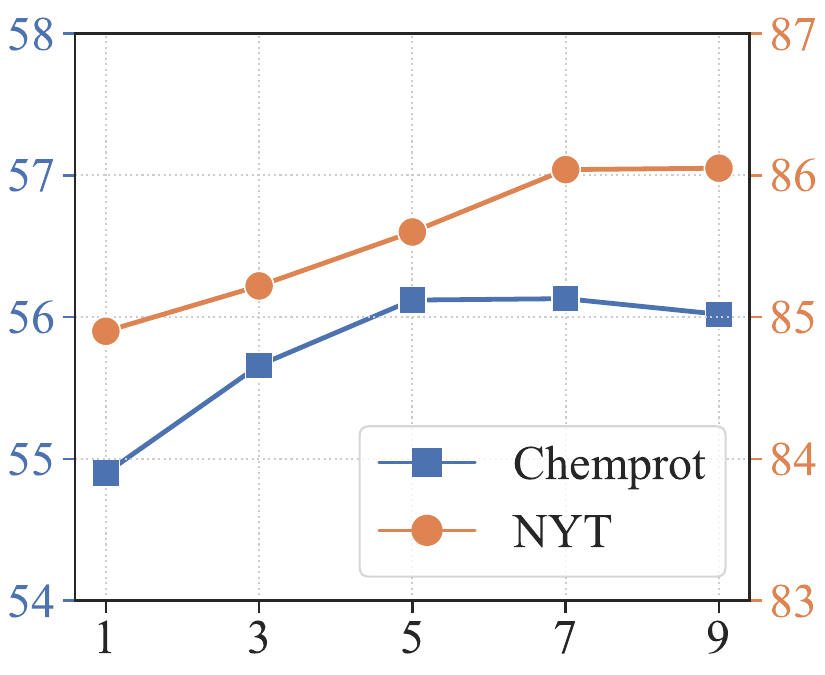}
		\label{fig:k}
	} \hspace{-1ex} %\hfill
	\subfloat[$c$]{
		\includegraphics[width=0.315\linewidth]{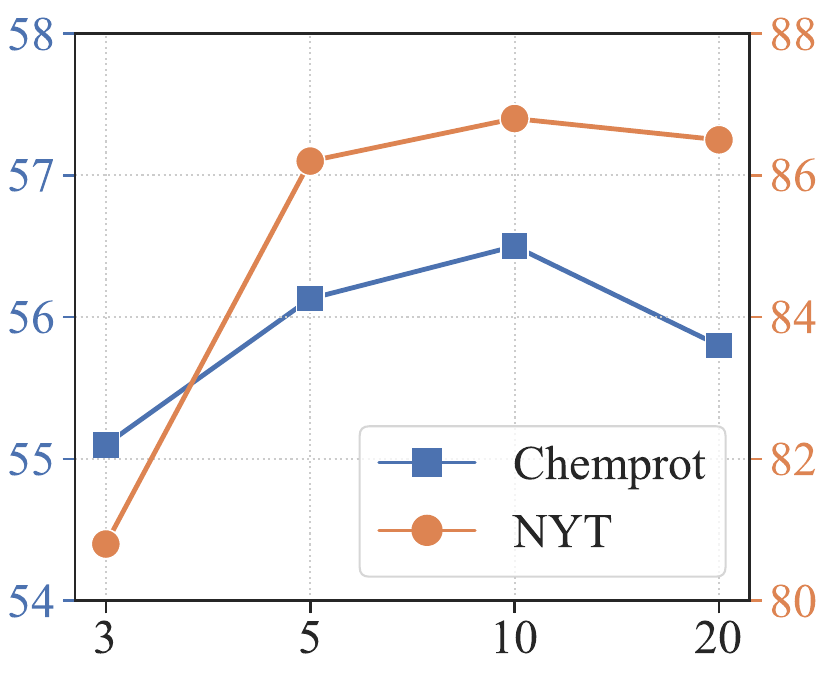}
		\label{fig:c}
	}
	\hspace{-1ex}
	\subfloat[Acc. over iters.]{
		\includegraphics[width=0.31\linewidth]{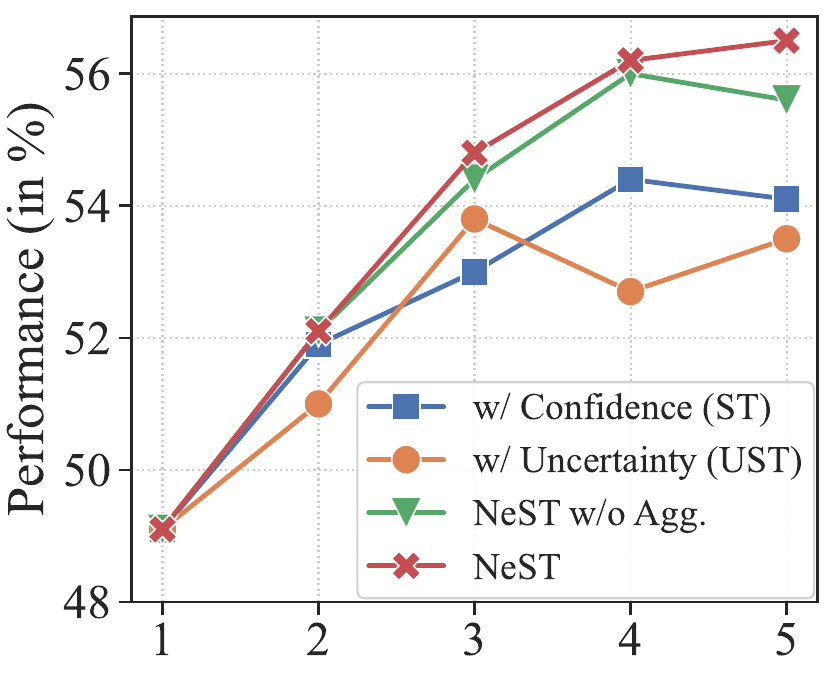}
		\label{fig:abl}
	} %\hfill
% 	\subfloat[Ablation]{
% 		\includegraphics[width=0.42\linewidth]{figures/ablation.pdf}
% 		\label{fig:abl}
% 	}
% 	\vspace{-2ex}
	\caption{Effect of different components of {\ours}.}
	\vspace{-1.5ex}
\label{fig:ablation}
\end{figure}

% \vspace{-0.5ex}
\subsection{Analysis}
% \vspace{-0.2ex}
\label{sec:case}
We take a closer look at the performance of {\ours} and other self-training algorithms using four text datasets. 
For each dataset, we study the setting with 30 labels per class.
% The analysis for graph datasets is in Appendix~\ref{apd:molecule}.

% molecule
\begin{figure}[t]
    \centering
     \vspace{-1ex}
    \includegraphics[width=0.93\linewidth]{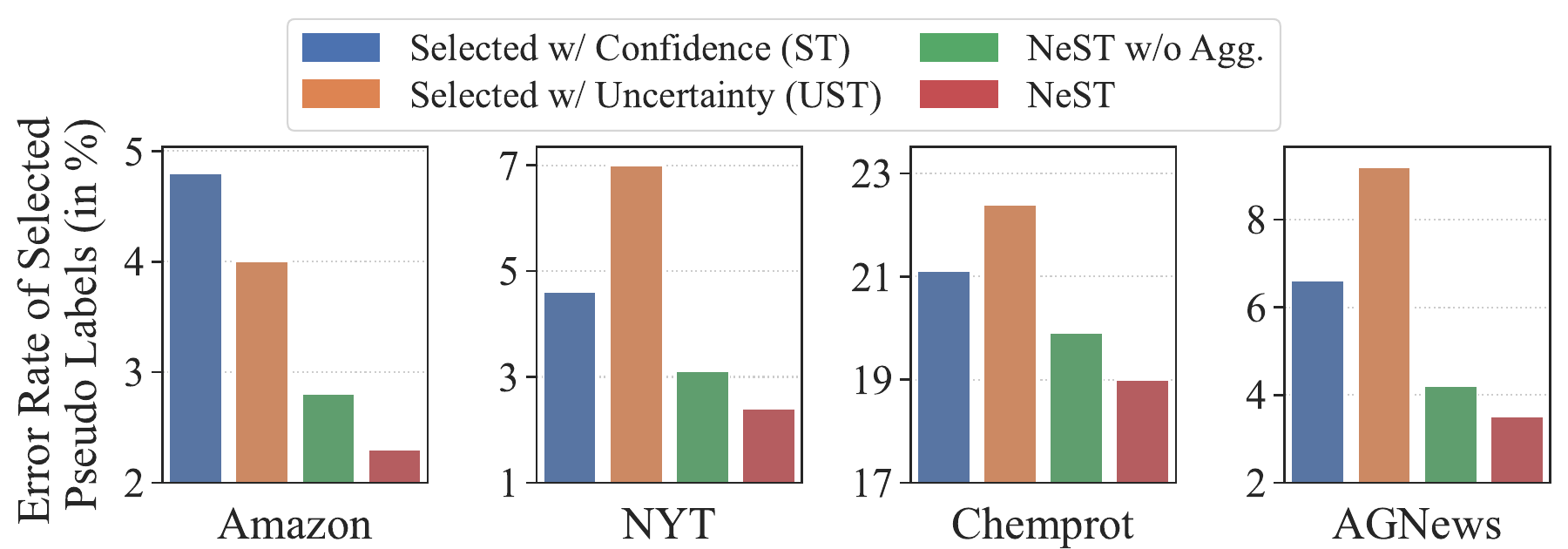}
    \caption{Error rates of pseudo labels selected by different methods. Agg. is the aggregation technique in Section~\ref{sec:selftrain}.}
    \label{fig:case}
    \vspace{-1ex}
\end{figure}

\begin{figure}[t]
    \centering
    \vspace{-1ex}
    \includegraphics[width=0.93\linewidth]{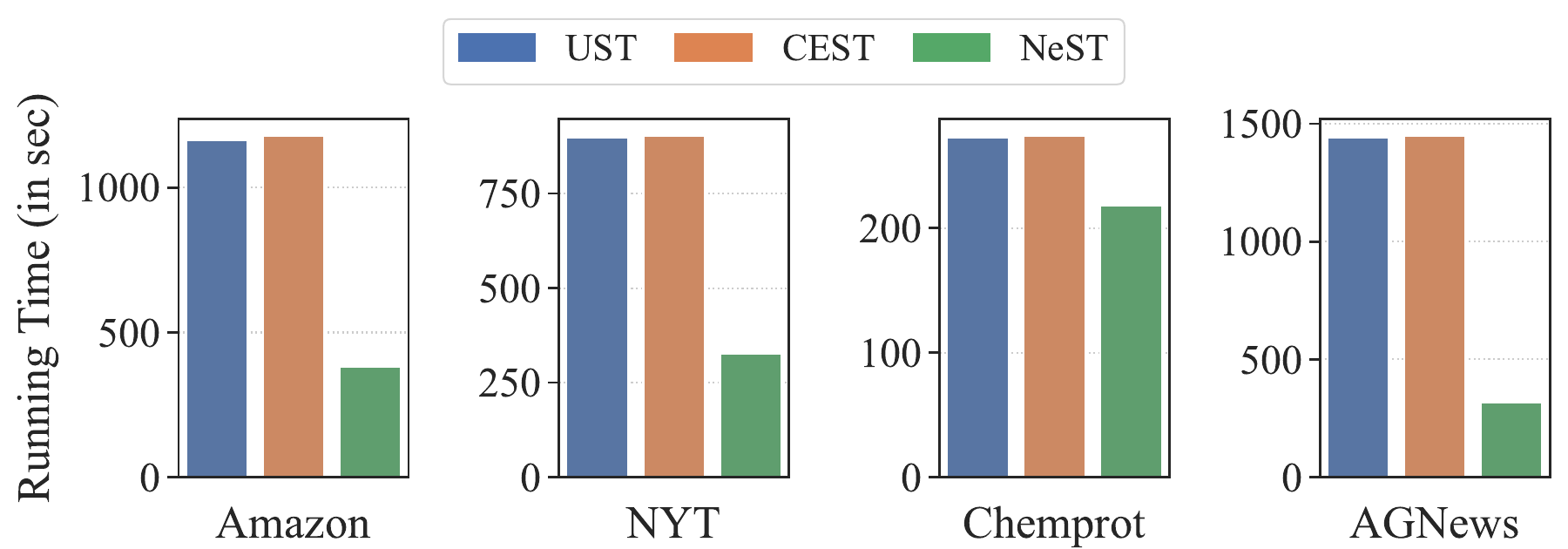}
    \caption{Running time of different methods.}
    \label{fig:runtime}
    \vspace{-2ex}
\end{figure}

\paragraph{Error of Pseudo Labels.}
% \label{sec:case}
To demonstrate how {\ours} reduces the noise of pseudo labels, we compare the error rate of pseudo labels selected by ST, UST and {\ours}.
From Figure~\ref{fig:case} we can notice that ST and UST tend to have high error rates due to their sole reliance on model prediction, and UST cannot stably improve the denoising ability of ST. In contrast, {\ours} significantly reduces the pseudo labels error rate by 36.8\% on average compared to the best baseline. As a result, cleaner pseudo labels lead to performance gain.
% \ran{Mention molecule dataset in appendix}

\paragraph{Running Time.}
We compare the running time for one self-training iteration of {\ours} with UST and CEST, which are the two strongest baselines using model uncertainty. 
As shown in Figure~\ref{fig:runtime}, the running time is reduced by 57.3\% on average. The gain is even more significant on larger datasets (e.g., AGNews) where inference multiple times becomes the efficiency bottleneck. 
Instead, the KNN operation takes less than 2 seconds with FAISS. To sum up, {\ours} is more efficient and can be readily combined with self-training. 

% \yue{hard label vs soft label}
% \yue{Effect of Noise-aware loss function, teacher/student model for prediction}
% \yue{Effect of $k$, $c$, $m$}

\section{Conclusion}
\label{sec:conclu}
% \vspace{-0.5ex}
In this paper, we propose {\ours} to improve sample selection in self-training for robust label efficient learning.
% Motivated by the fact that data with the same class share similar representation, 
We design a neighborhood-regularized approach to select more reliable samples based on representations for self-training. Moreover, we propose to aggregate the predictions on different iterations to stabilize self-training. 
Experiments on four text datasets and four graph datasets show that {\ours} outperforms the baselines by 1.83\% and 2.51\% on average.  
{\ours} also significantly reduce the noise in pseudo labels by 36.8\% and reduce the running time by 57.3\% when compared with the strongest baseline. 
For future works, we plan to extend {\ours} to other application domains and modalities. 

% \newpage 

\section*{Acknowledgement}
We thank the anonymous reviewers for the valuable feedbacks. 
This research was partially supported by the internal funds and GPU servers provided by the Computer Science Department of Emory University. 
In addition, YY and CZ were partly supported by NSF IIS-2008334, IIS-2106961, and CAREER IIS-2144338.
JH was supported by NSF grants IIS-1838200 and IIS-2145411. 

% Use \bibliography{yourbibfile} instead or the References section will not appear in your paper
% \nobibliography{aaai23}
\bibliography{aaai23,ref}

\begin{thebibliography}{69}
\providecommand{\natexlab}[1]{#1}

\bibitem[{Arazo et~al.(2020)Arazo, Ortego, Albert, O’Connor, and
  McGuinness}]{arazo2020pseudo}
Arazo, E.; Ortego, D.; Albert, P.; O’Connor, N.~E.; and McGuinness, K. 2020.
\newblock Pseudo-labeling and confirmation bias in deep semi-supervised
  learning.
\newblock In \emph{IJCNN}.

\bibitem[{Bragg et~al.(2021)Bragg, Cohan, Lo, and Beltagy}]{bragg2021flex}
Bragg, J.; Cohan, A.; Lo, K.; and Beltagy, I. 2021.
\newblock Flex: Unifying evaluation for few-shot nlp.
\newblock \emph{NeurIPS}.

\bibitem[{Chen, Yang, and Yang(2020)}]{chen2020mixtext}
Chen, J.; Yang, Z.; and Yang, D. 2020.
\newblock Mixtext: Linguistically-informed interpolation of hidden space for
  semi-supervised text classification.
\newblock In \emph{ACL}.

\bibitem[{Chen et~al.(2020)Chen, Kornblith, Norouzi, and
  Hinton}]{chen2020simple}
Chen, T.; Kornblith, S.; Norouzi, M.; and Hinton, G. 2020.
\newblock A simple framework for contrastive learning of visual
  representations.
\newblock In \emph{ICML}.

\bibitem[{Cohen, Mori-S{\'a}nchez, and Yang(2012)}]{cohen2012challenges}
Cohen, A.~J.; Mori-S{\'a}nchez, P.; and Yang, W. 2012.
\newblock Challenges for density functional theory.
\newblock \emph{Chemical reviews}.

\bibitem[{Cui et~al.(2022)Cui, Lu, Ge, and Yang}]{cui2022can}
Cui, H.; Lu, J.; Ge, Y.; and Yang, C. 2022.
\newblock How Can Graph Neural Networks Help Document Retrieval: A Case Study
  on CORD19 with Concept Map Generation.
\newblock In \emph{ECIR}.

\bibitem[{Delaney(2004)}]{delaney2004esol}
Delaney, J.~S. 2004.
\newblock ESOL: estimating aqueous solubility directly from molecular
  structure.
\newblock \emph{J. Chem. Inf. Comp. Sci.}

\bibitem[{Devlin et~al.(2019)Devlin, Chang, Lee, and
  Toutanova}]{devlin2018bert}
Devlin, J.; Chang, M.-W.; Lee, K.; and Toutanova, K. 2019.
\newblock Bert: Pre-training of deep bidirectional transformers for language
  understanding.
\newblock In \emph{NAACL-HLT}.

\bibitem[{Du et~al.(2020)Du, Grave, Gunel, Chaudhary, Celebi, Auli, Stoyanov,
  and Conneau}]{du2020self}
Du, J.; Grave, E.; Gunel, B.; Chaudhary, V.; Celebi, O.; Auli, M.; Stoyanov,
  V.; and Conneau, A. 2020.
\newblock Self-training improves pre-training for natural language
  understanding.
\newblock \emph{arXiv preprint arXiv:2010.02194}.

\bibitem[{Gal and Ghahramani(2016)}]{gal2016dropout}
Gal, Y.; and Ghahramani, Z. 2016.
\newblock Dropout as a bayesian approximation: Representing model uncertainty
  in deep learning.
\newblock In \emph{ICML}.

\bibitem[{Gaulton et~al.(2012)Gaulton, Bellis, Bento, Chambers, Davies, Hersey,
  Light, McGlinchey, Michalovich, Al-Lazikani et~al.}]{lipo}
Gaulton, A.; Bellis, L.~J.; Bento, A.~P.; Chambers, J.; Davies, M.; Hersey, A.;
  Light, Y.; McGlinchey, S.; Michalovich, D.; Al-Lazikani, B.; et~al. 2012.
\newblock ChEMBL: a large-scale bioactivity database for drug discovery.
\newblock \emph{Nucleic acids research}.

\bibitem[{Guo et~al.(2017)Guo, Pleiss, Sun, and
  Weinberger}]{guo2017calibration}
Guo, C.; Pleiss, G.; Sun, Y.; and Weinberger, K.~Q. 2017.
\newblock On calibration of modern neural networks.
\newblock In \emph{ICML}.

\bibitem[{Gururangan et~al.(2019)Gururangan, Dang, Card, and
  Smith}]{gururangan2019variational}
Gururangan, S.; Dang, T.; Card, D.; and Smith, N.~A. 2019.
\newblock Variational Pretraining for Semi-supervised Text Classification.
\newblock In \emph{ACL}.

\bibitem[{Hao et~al.(2020)Hao, Lu, Huang, Wang, Hu, Liu, Chen, and
  Lee}]{hao2020asgn}
Hao, Z.; Lu, C.; Huang, Z.; Wang, H.; Hu, Z.; Liu, Q.; Chen, E.; and Lee, C.
  2020.
\newblock ASGN: An active semi-supervised graph neural network for molecular
  property prediction.
\newblock In \emph{KDD}.

\bibitem[{Hu et~al.(2020)Hu, Liu, Gomes, Zitnik, Liang, Pande, and
  Leskovec}]{Hu2020Strategies}
Hu, W.; Liu, B.; Gomes, J.; Zitnik, M.; Liang, P.; Pande, V.; and Leskovec, J.
  2020.
\newblock Strategies for Pre-training Graph Neural Networks.
\newblock In \emph{ICLR}.

\bibitem[{Johnson, Douze, and Jégou(2021)}]{faiss}
Johnson, J.; Douze, M.; and Jégou, H. 2021.
\newblock Billion-Scale Similarity Search with GPUs.
\newblock \emph{IEEE TBD}.

\bibitem[{Kan, Cui, and Yang(2021)}]{kan2021zero}
Kan, X.; Cui, H.; and Yang, C. 2021.
\newblock Zero-shot scene graph relation prediction through commonsense
  knowledge integration.
\newblock In \emph{ECML-PKDD}.

\bibitem[{Kim et~al.(2022{\natexlab{a}})Kim, Woo, Oh, Cha, and
  Han}]{kim2022alp}
Kim, H.~H.; Woo, D.; Oh, S.~J.; Cha, J.-W.; and Han, Y.-S. 2022{\natexlab{a}}.
\newblock ALP: Data Augmentation using Lexicalized PCFGs for Few-Shot Text
  Classification.
\newblock In \emph{AAAI}.

\bibitem[{Kim et~al.(2022{\natexlab{b}})Kim, Min, Kim, Lee, Seo, Ryoo, and
  Kim}]{kim2022conmatch}
Kim, J.; Min, Y.; Kim, D.; Lee, G.; Seo, J.; Ryoo, K.; and Kim, S.
  2022{\natexlab{b}}.
\newblock ConMatch: Semi-supervised Learning with Confidence-Guided Consistency
  Regularization.
\newblock In \emph{ECCV}.

\bibitem[{Kingma and Ba(2014)}]{kingma2014adam}
Kingma, D.~P.; and Ba, J. 2014.
\newblock Adam: A method for stochastic optimization.
\newblock \emph{arXiv preprint arXiv:1412.6980}.

\bibitem[{Kong et~al.(2020)Kong, Jiang, Zhuang, Lyu, Zhao, and
  Zhang}]{kong2020calibrated}
Kong, L.; Jiang, H.; Zhuang, Y.; Lyu, J.; Zhao, T.; and Zhang, C. 2020.
\newblock Calibrated Language Model Fine-Tuning for In-and Out-of-Distribution
  Data.
\newblock In \emph{EMNLP}.

\bibitem[{Lang, Vijayaraghavan, and Sontag(2022)}]{lang2022training}
Lang, H.; Vijayaraghavan, A.; and Sontag, D. 2022.
\newblock Training Subset Selection for Weak Supervision.
\newblock In \emph{NeurIPS}.

\bibitem[{Lee et~al.(2020)Lee, Yoon, Kim, Kim, Kim, So, and
  Kang}]{lee2020biobert}
Lee, J.; Yoon, W.; Kim, S.; Kim, D.; Kim, S.; So, C.~H.; and Kang, J. 2020.
\newblock BioBERT: a pre-trained biomedical language representation model for
  biomedical text mining.
\newblock \emph{Bioinformatics}.

\bibitem[{Li, Savarese, and Hoi(2022)}]{li2022masked}
Li, J.; Savarese, S.; and Hoi, S.~C. 2022.
\newblock Masked Unsupervised Self-training for Zero-shot Image Classification.
\newblock \emph{arXiv preprint arXiv:2206.02967}.

\bibitem[{Li, Xiong, and Hoi(2021)}]{li2021comatch}
Li, J.; Xiong, C.; and Hoi, S.~C. 2021.
\newblock Comatch: Semi-supervised learning with contrastive graph
  regularization.
\newblock In \emph{ICCV}.

\bibitem[{Li et~al.(2021)Li, Zhang, Xu, Dickerson, and Ba}]{li2021how}
Li, J.; Zhang, M.; Xu, K.; Dickerson, J.~P.; and Ba, J. 2021.
\newblock How does a Neural Network's Architecture Impact its Robustness to
  Noisy Labels?
\newblock In \emph{NeurIPS}.

\bibitem[{Liang et~al.(2020)Liang, Yu, Jiang, Er, Wang, Zhao, and
  Zhang}]{liang2020bond}
Liang, C.; Yu, Y.; Jiang, H.; Er, S.; Wang, R.; Zhao, T.; and Zhang, C. 2020.
\newblock Bond: Bert-assisted open-domain named entity recognition with distant
  supervision.
\newblock In \emph{KDD}.

\bibitem[{Liu et~al.(2022)Liu, Tian, Chen, Liu, Belagiannis, and
  Carneiro}]{liu2022acpl}
Liu, F.; Tian, Y.; Chen, Y.; Liu, Y.; Belagiannis, V.; and Carneiro, G. 2022.
\newblock ACPL: Anti-Curriculum Pseudo-Labelling for Semi-Supervised Medical
  Image Classification.
\newblock In \emph{CVPR}, 20697--20706.

\bibitem[{Liu et~al.(2019)Liu, Ott, Goyal, Du, Joshi, Chen, Levy, Lewis,
  Zettlemoyer, and Stoyanov}]{liu2019roberta}
Liu, Y.; Ott, M.; Goyal, N.; Du, J.; Joshi, M.; Chen, D.; Levy, O.; Lewis, M.;
  Zettlemoyer, L.; and Stoyanov, V. 2019.
\newblock Roberta: A robustly optimized bert pretraining approach.
\newblock \emph{arXiv preprint arXiv:1907.11692}.

\bibitem[{Martins et~al.(2012)Martins, Teixeira, Pinheiro, and Falcao}]{bbbp}
Martins, I.~F.; Teixeira, A.~L.; Pinheiro, L.; and Falcao, A.~O. 2012.
\newblock A Bayesian approach to in silico blood-brain barrier penetration
  modeling.
\newblock \emph{J. Chem. Inf. Model.}

\bibitem[{McAuley and Leskovec(2013)}]{mcauley2013hidden}
McAuley, J.; and Leskovec, J. 2013.
\newblock Hidden factors and hidden topics: understanding rating dimensions
  with review text.
\newblock In \emph{RecSys}.

\bibitem[{Meng et~al.(2020)Meng, Huang, Wang, Wang, Zhang, Zhang, and
  Han}]{meng2020discriminative}
Meng, Y.; Huang, J.; Wang, G.; Wang, Z.; Zhang, C.; Zhang, Y.; and Han, J.
  2020.
\newblock Discriminative topic mining via category-name guided text embedding.
\newblock In \emph{WWW}.

\bibitem[{Menon et~al.(2021)Menon, Jayasumana, Rawat, Jain, Veit, and
  Kumar}]{menon2021longtail}
Menon, A.~K.; Jayasumana, S.; Rawat, A.~S.; Jain, H.; Veit, A.; and Kumar, S.
  2021.
\newblock Long-tail learning via logit adjustment.
\newblock In \emph{ICLR}.

\bibitem[{Miyato et~al.(2018)Miyato, Maeda, Koyama, and
  Ishii}]{miyato2018virtual}
Miyato, T.; Maeda, S.-i.; Koyama, M.; and Ishii, S. 2018.
\newblock Virtual adversarial training: a regularization method for supervised
  and semi-supervised learning.
\newblock \emph{TPAMI}.

\bibitem[{Mukherjee and Awadallah(2020)}]{mukherjee2020uncertainty}
Mukherjee, S.; and Awadallah, A. 2020.
\newblock Uncertainty-aware self-training for few-shot text classification.
\newblock \emph{NeurIPS}.

\bibitem[{Rizve et~al.(2021)Rizve, Duarte, Rawat, and Shah}]{rizve2021in}
Rizve, M.~N.; Duarte, K.; Rawat, Y.~S.; and Shah, M. 2021.
\newblock In Defense of Pseudo-Labeling: An Uncertainty-Aware Pseudo-label
  Selection Framework for Semi-Supervised Learning.
\newblock In \emph{ICLR}.

\bibitem[{Rong et~al.(2020)Rong, Bian, Xu, Xie, Wei, Huang, and
  Huang}]{rong2020grover}
Rong, Y.; Bian, Y.; Xu, T.; Xie, W.; Wei, Y.; Huang, W.; and Huang, J. 2020.
\newblock Grover: Self-supervised message passing transformer on large-scale
  molecular data.
\newblock \emph{NeurIPS}.

\bibitem[{Rosenberg, Hebert, and Schneiderman(2005)}]{rosenberg2005semi}
Rosenberg, C.; Hebert, M.; and Schneiderman, H. 2005.
\newblock Semi-Supervised Self-Training of Object Detection Models.
\newblock In \emph{WACV/MOTION}.

\bibitem[{Sohn et~al.(2020)Sohn, Berthelot, Carlini, Zhang, Zhang, Raffel,
  Cubuk, Kurakin, and Li}]{sohn2020fixmatch}
Sohn, K.; Berthelot, D.; Carlini, N.; Zhang, Z.; Zhang, H.; Raffel, C.~A.;
  Cubuk, E.~D.; Kurakin, A.; and Li, C.-L. 2020.
\newblock Fixmatch: Simplifying semi-supervised learning with consistency and
  confidence.
\newblock \emph{NeurIPS}.

\bibitem[{Subramanian et~al.(2016)Subramanian, Ramsundar, Pande, and
  Denny}]{bace}
Subramanian, G.; Ramsundar, B.; Pande, V.; and Denny, R.~A. 2016.
\newblock Computational modeling of $\beta$-secretase 1 (BACE-1) inhibitors
  using ligand based approaches.
\newblock \emph{J. Chem. Inf. Model.}

\bibitem[{Sun et~al.(2020)Sun, Hoffman, Verma, and Tang}]{sun2020infograph}
Sun, F.-Y.; Hoffman, J.; Verma, V.; and Tang, J. 2020.
\newblock InfoGraph: Unsupervised and Semi-supervised Graph-Level
  Representation Learning via Mutual Information Maximization.
\newblock In \emph{ICLR}.

\bibitem[{Taboureau et~al.(2010)Taboureau, Nielsen, Audouze, Weinhold,
  Edsg{\"a}rd, Roque, Kouskoumvekaki, Bora, Curpan, Jensen
  et~al.}]{taboureau2010chemprot}
Taboureau, O.; Nielsen, S.~K.; Audouze, K.; Weinhold, N.; Edsg{\"a}rd, D.;
  Roque, F.~S.; Kouskoumvekaki, I.; Bora, A.; Curpan, R.; Jensen, T.~S.; et~al.
  2010.
\newblock ChemProt: a disease chemical biology database.
\newblock \emph{Nucleic acids research}.

\bibitem[{Tarvainen and Valpola(2017)}]{tarvainen2017mean}
Tarvainen, A.; and Valpola, H. 2017.
\newblock Mean teachers are better role models: Weight-averaged consistency
  targets improve semi-supervised deep learning results.
\newblock \emph{NeurIPS}.

\bibitem[{Tsai, Lin, and Fu(2022)}]{tsai2022contrast}
Tsai, A. C.-Y.; Lin, S.-Y.; and Fu, L.-C. 2022.
\newblock Contrast-enhanced Semi-supervised Text Classification with Few
  Labels.
\newblock In \emph{AAAI}.

\bibitem[{Wang et~al.(2022)Wang, Chen, Fan, Wang, Tao, Hou, Wang, Yang, Zhou,
  Guo et~al.}]{wang2022usb}
Wang, Y.; Chen, H.; Fan, Y.; Wang, S.; Tao, R.; Hou, W.; Wang, R.; Yang, L.;
  Zhou, Z.; Guo, L.-Z.; et~al. 2022.
\newblock {USB}: A Unified Semi-supervised Learning Benchmark for
  Classification.
\newblock In \emph{NeurIPS Datasets and Benchmarks Track}.

\bibitem[{Wang et~al.(2021)Wang, Mukherjee, Chu, Tu, Wu, Gao, and
  Awadallah}]{wang2021meta}
Wang, Y.; Mukherjee, S.; Chu, H.; Tu, Y.; Wu, M.; Gao, J.; and Awadallah, A.~H.
  2021.
\newblock Meta Self-training for Few-shot Neural Sequence Labeling.
\newblock In \emph{KDD}.

\bibitem[{Wolf et~al.(2019)Wolf, Debut, Sanh, Chaumond, Delangue, Moi, Cistac,
  Rault, Louf, Funtowicz et~al.}]{wolf2019huggingface}
Wolf, T.; Debut, L.; Sanh, V.; Chaumond, J.; Delangue, C.; Moi, A.; Cistac, P.;
  Rault, T.; Louf, R.; Funtowicz, M.; et~al. 2019.
\newblock HuggingFace's Transformers: State-of-the-art natural language
  processing.
\newblock \emph{arXiv preprint arXiv:1910.03771}.

\bibitem[{Wu et~al.(2018)Wu, Ramsundar, Feinberg, Gomes, Geniesse, Pappu,
  Leswing, and Pande}]{wu2018moleculenet}
Wu, Z.; Ramsundar, B.; Feinberg, E.~N.; Gomes, J.; Geniesse, C.; Pappu, A.~S.;
  Leswing, K.; and Pande, V. 2018.
\newblock MoleculeNet: a benchmark for molecular machine learning.
\newblock \emph{Chemical science}, 9(2).

\bibitem[{Xia et~al.(2022)Xia, Liu, Han, Gong, Yu, Niu, and
  Sugiyama}]{xia2022sample}
Xia, X.; Liu, T.; Han, B.; Gong, M.; Yu, J.; Niu, G.; and Sugiyama, M. 2022.
\newblock Sample Selection with Uncertainty of Losses for Learning with Noisy
  Labels.
\newblock In \emph{ICLR}.

\bibitem[{Xie et~al.(2020{\natexlab{a}})Xie, Dai, Hovy, Luong, and
  Le}]{xie2020uda}
Xie, Q.; Dai, Z.; Hovy, E.; Luong, M.-T.; and Le, Q.~V. 2020{\natexlab{a}}.
\newblock Unsupervised Data Augmentation for Consistency Training.
\newblock In \emph{NeurIPS}.

\bibitem[{Xie et~al.(2020{\natexlab{b}})Xie, Luong, Hovy, and Le}]{xie2020self}
Xie, Q.; Luong, M.-T.; Hovy, E.; and Le, Q.~V. 2020{\natexlab{b}}.
\newblock Self-training with noisy student improves imagenet classification.
\newblock In \emph{CVPR}.

\bibitem[{Xiong et~al.(2019)Xiong, Wang, Liu, Zhong, Wan, Li, Li, Luo, Chen
  et~al.}]{xiong2019pushing}
Xiong, Z.; Wang, D.; Liu, X.; Zhong, F.; Wan, X.; Li, X.; Li, Z.; Luo, X.;
  Chen, K.; et~al. 2019.
\newblock Pushing the boundaries of molecular representation for drug discovery
  with the graph attention mechanism.
\newblock \emph{Journal of medicinal chemistry}.

\bibitem[{Xu et~al.(2021)Xu, Ding, Zhang, and Zhou}]{xu2021dp}
Xu, Y.; Ding, J.; Zhang, L.; and Zhou, S. 2021.
\newblock DP-SSL: Towards Robust Semi-supervised Learning with A Few Labeled
  Samples.
\newblock \emph{NeurIPS}.

\bibitem[{Yang et~al.(2021)Yang, Song, King, and Xu}]{yang2021survey}
Yang, X.; Song, Z.; King, I.; and Xu, Z. 2021.
\newblock A survey on deep semi-supervised learning.
\newblock \emph{arXiv preprint arXiv:2103.00550}.

\bibitem[{You et~al.(2021)You, Chen, Shen, and Wang}]{you21a}
You, Y.; Chen, T.; Shen, Y.; and Wang, Z. 2021.
\newblock Graph Contrastive Learning Automated.
\newblock In \emph{ICML}.

\bibitem[{Yu et~al.(2022{\natexlab{a}})Yu, Kong, Zhang, Zhang, and
  Zhang}]{yu2022actune}
Yu, Y.; Kong, L.; Zhang, J.; Zhang, R.; and Zhang, C. 2022{\natexlab{a}}.
\newblock AcTune: Uncertainty-Based Active Self-Training for Active Fine-Tuning
  of Pretrained Language Models.
\newblock In \emph{NAACL}.

\bibitem[{Yu et~al.(2022{\natexlab{b}})Yu, Zhang, Xu, Zhang, Shen, and
  Zhang}]{yu2022cold}
Yu, Y.; Zhang, R.; Xu, R.; Zhang, J.; Shen, J.; and Zhang, C.
  2022{\natexlab{b}}.
\newblock Cold-Start Data Selection for Few-shot Language Model Fine-tuning: A
  Prompt-Based Uncertainty Propagation Approach.
\newblock \emph{arXiv preprint arXiv:2209.06995}.

\bibitem[{Yu et~al.(2021)Yu, Zuo, Jiang, Ren, Zhao, and Zhang}]{yu2021fine}
Yu, Y.; Zuo, S.; Jiang, H.; Ren, W.; Zhao, T.; and Zhang, C. 2021.
\newblock Fine-Tuning Pre-trained Language Model with Weak Supervision: A
  Contrastive-Regularized Self-Training Approach.
\newblock In \emph{NAACL}.

\bibitem[{Zhang et~al.(2021{\natexlab{a}})Zhang, Wang, Hou, Wu, Wang, Okumura,
  and Shinozaki}]{zhang2021flexmatch}
Zhang, B.; Wang, Y.; Hou, W.; Wu, H.; Wang, J.; Okumura, M.; and Shinozaki, T.
  2021{\natexlab{a}}.
\newblock Flexmatch: Boosting semi-supervised learning with curriculum pseudo
  labeling.
\newblock \emph{NeurIPS}.

\bibitem[{Zhang et~al.(2021{\natexlab{b}})Zhang, Yu, Li, Wang, Yang, Yang, and
  Ratner}]{zhang2021wrench}
Zhang, J.; Yu, Y.; Li, Y.; Wang, Y.; Yang, Y.; Yang, M.; and Ratner, A.
  2021{\natexlab{b}}.
\newblock WRENCH: A Comprehensive Benchmark for Weak Supervision.
\newblock In \emph{NeurIPS Datasets and Benchmarks Track}.

\bibitem[{Zhang, Yu, and Zhang(2020)}]{zhang2020seqmix}
Zhang, R.; Yu, Y.; and Zhang, C. 2020.
\newblock Seqmix: Augmenting active sequence labeling via sequence mixup.
\newblock \emph{EMNLP}.

\bibitem[{Zhang et~al.(2020)Zhang, Hu, Subramonian, and Sun}]{zhang2020motif}
Zhang, S.; Hu, Z.; Subramonian, A.; and Sun, Y. 2020.
\newblock Motif-driven contrastive learning of graph representations.
\newblock \emph{arXiv preprint arXiv:2012.12533}.

\bibitem[{Zhang, Zhao, and LeCun(2015)}]{zhang2015character}
Zhang, X.; Zhao, J.; and LeCun, Y. 2015.
\newblock Character-level Convolutional Networks for Text Classification.
\newblock In \emph{NeurIPS}.

\bibitem[{Zhao et~al.(2022)Zhao, Zhou, Wang, Shi, and Gao}]{zhao2022lassl}
Zhao, Z.; Zhou, L.; Wang, L.; Shi, Y.; and Gao, Y. 2022.
\newblock LaSSL: Label-guided Self-training for Semi-supervised Learning.
\newblock In \emph{AAAI}.

\bibitem[{Zheng et~al.(2022)Zheng, You, Huang, Wang, Qian, and
  Xu}]{zheng2022simmatch}
Zheng, M.; You, S.; Huang, L.; Wang, F.; Qian, C.; and Xu, C. 2022.
\newblock SimMatch: Semi-supervised Learning with Similarity Matching.
\newblock In \emph{CVPR}.

\bibitem[{Zhou, Kantarcioglu, and Thuraisingham(2012)}]{zhou2012self}
Zhou, Y.; Kantarcioglu, M.; and Thuraisingham, B. 2012.
\newblock Self-training with selection-by-rejection.
\newblock In \emph{ICDM}.

\bibitem[{Zhu et~al.(2022)Zhu, Xu, Cui et~al.}]{zhu2022structure}
Zhu, Y.; Xu, Y.; Cui, H.; et~al. 2022.
\newblock Structure-enhanced heterogeneous graph contrastive learning.
\newblock In \emph{SDM}.

\bibitem[{Zhu et~al.(2021)Zhu, Xu, Yu, Liu, Wu, and Wang}]{zhu2021graph}
Zhu, Y.; Xu, Y.; Yu, F.; Liu, Q.; Wu, S.; and Wang, L. 2021.
\newblock Graph contrastive learning with adaptive augmentation.
\newblock In \emph{WWW}.

\bibitem[{Zhu, Dong, and Liu(2022)}]{zhu2022detecting}
Zhu, Z.; Dong, Z.; and Liu, Y. 2022.
\newblock Detecting corrupted labels without training a model to predict.
\newblock In \emph{ICML}.

\end{thebibliography}

\appendix
% \clearpage
\section{Dataset Information}
\label{apd:first}
\begin{itemize}
    \item \textbf{AGNews}~\cite{zhang2015character}: This dataset is a collection of more than one million news articles. It is constructed by \citet{zhang2015character} choosing the 4 largest topic classes from the original corpus. The total number of training samples is 120K and both validation and testing are 7.6K.
    \item \textbf{Elec}~\cite{mcauley2013hidden}: This dataset is a subset of Amazon's product reviews for binary sentiment classification. It is constructed by \citet{mcauley2013hidden}, including 25K training samples,  and 25K testing samples.
    \item \textbf{NYT}~\cite{meng2020discriminative}: This dataset is collected using the New York Times API. It is constructed by \citet{meng2020discriminative}, including 30K training samples,  and 3K testing samples.
    \item \textbf{ChemProt}~\cite{taboureau2010chemprot}. This is a 10-class relation extraction dataset constructed by \citet{taboureau2010chemprot}, containing 12K training samples and 1.6K testing samples.
    \item \textbf{BBBP}~\cite{bbbp}: This is a binary classification task for predicting whether a compound carries the permeability property of penetrating the blood-brain barrier. It contains 2039 molecules in total.
    \item \textbf{BACE}~\cite{bace}: This is a binary classification task for predicting compounds which could act as the inhibitors of human $\beta$-secretase 1 in the past few years. It contains 1513 molecules in total.
    \item \textbf{Esol}~\cite{delaney2004esol} is a regression task for predicting the solubility of compounds. It contains 1128 molecules in total. 
    \item \textbf{Lipophilicity}~\cite{lipo} is a regression task for predicting the property that affects the molecular membrane permeability and solubility. It contains 4200 molecules in total. 
\end{itemize}

\section{Introduction of Confidence-based and Uncertainty-based Selection Criterion}
\label{apd:select}
\subsection{Confidence-based Method}
For confidence-based strategy, we grant higher selection probability $\by_i = p(x_i)$ for sample $x_i$ with higher predictive confidence 
\begin{equation}
p(x_i) \propto \frac{\max\left({\by_i}\right)}{\sum_{x_{u} \in \cX_{u}} \max\left({\by_u}\right)}
\end{equation}

\subsection{Uncertainty-based Method}
% \cite{cascante2021curriculum}
For uncertainty-based method, the overall goal is to select samples that model is less uncertain about. 
For each data sample $x_i \in \cX_u$ , the information gain $\mathbb{B}$ with respect to its expected label $y_i$ as 
\begin{equation}
\mathbb{B}\left(y_i, \theta  \mid x_i, D_{U}\right)=\mathbb{H}\left(y \mid x, \cX_{u}\right)-\mathbb{E}_{p\left(\theta \mid D_{U}\right)}[\mathbb{H}(y \mid x, \theta)]
\end{equation}

As the above equation is computationally intractable, BALD~\cite{gal2016dropout} is used to approximate the model uncertainty as
\begin{small}
\begin{align}
    \hat{\mathbb{B}}\left(y_{i}, \theta \mid x_{i}, \cX_{u}\right)&=-\sum_{c=1}^{C}\left(\frac{1}{M} \sum_{m=1}^{M} {\by}_{c}^{m}\right) \log \left(\frac{1}{M} \sum_{m=1}^{M} {\by}_{c}^{m}\right) \nonumber \\ 
    &+ \frac{1}{M} \sum_{c=1}^{C}\sum_{m=1}^{M} {\by}_{c}^{m} \log ({\by}_{c}^{m})
\end{align}
\end{small}

Then, the sample selection probability is calculated as 
\begin{equation}
p(x_i) \propto \frac{1-\widehat{\mathbb{B}}\left({\by}_{i} \mid x_{i}, \cX_{u}\right)}{\sum_{x_{u} \in \cX_{u}} 1-\widehat{\mathbb{B}}\left(\by_{u} \mid x_{u}, \cX_{u}\right)}.
\end{equation}

Note that, in the above equations, $M$ denotes the number of inferences. Usually $M$ is set to a large value ($M=10$ is used in this work). This causes the inefficiency issue for deep neural networks, especially for large datasets.  

\section{Implementations}
\subsection{Computing Infrastructure}
\textbf{System}: Ubuntu 18.04.3 LTS; Python 3.7; Pytorch 1.8. \\
\textbf{CPU}: Intel(R) Core(TM) i7-5930K CPU @ 3.50GHz. \\
\textbf{GPU}: NVIDIA A5000. \\

\subsection{Hyperparameters}
\label{apd:param}
Details of hyperparameters in text datasets is shown in Table~\ref{tab:hyperparameter-semi-text}. 
Details of hyperparameters in graph datasets is shown in Table~\ref{tab:hyperparameter-semi-graph}.

\begin{table*}
\centering
% \vskip -0.1in
\begin{tabular}{ccccc}
\toprule 
\bf Hyper-parameter &\bf AGNews& \bf  Elec & \bf NYT & \bf Chemprot  \\ \midrule 
Dropout Ratio & \multicolumn{4}{@{\hskip1pt}c@{\hskip1pt}}{0.1}  \\ 
Maximum Tokens  & 128 & 256 & 160 & 64 \\ 
Batch Size for Labeled Data  & 16 & 8 & 16 & 32 \\ 
Batch Size for Unlabeled Data  & \multicolumn{4}{@{\hskip1pt}c@{\hskip1pt}}{32} \\ 
Learning Rate & 2e-5 & 2e-5 & 1e-5 & 5e-5   \\
% Weight Decay & 1e-6 \\ \hline
Initialization Epochs & 12 & 15 & 15 & 15  \\ 
% $T$ &  4000 & 3000 & 4000  \\ \hline
$k$ & 5 & 7 & 9 & 7    \\ 
$c$ & 10 & 20 & 10 & 10  \\ 
\bottomrule
\end{tabular}
\caption{Hyper-parameter configurations for semi-supervised text classification.}
\label{tab:hyperparameter-semi-text}
\end{table*}

\section{Hyperparameter Study on Graph Datasets}
\label{apd:abla}
\begin{figure}
	\centering
	\vspace{-3ex}
	\subfloat[$k$]{
		\includegraphics[width=0.45\linewidth]{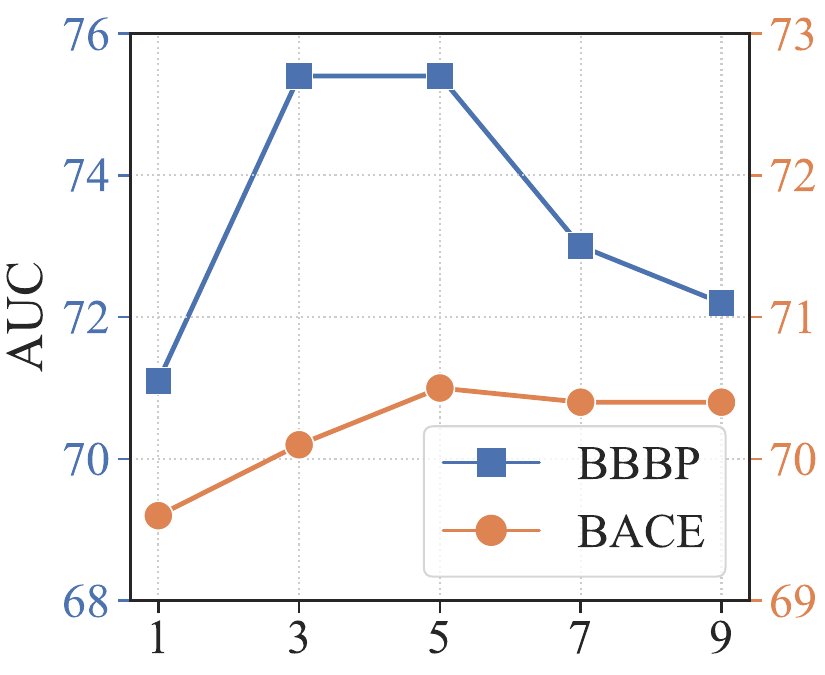}
		\label{fig:k}
	} \hspace{-1ex} %\hfill
	\subfloat[$c$]{
		\includegraphics[width=0.45\linewidth]{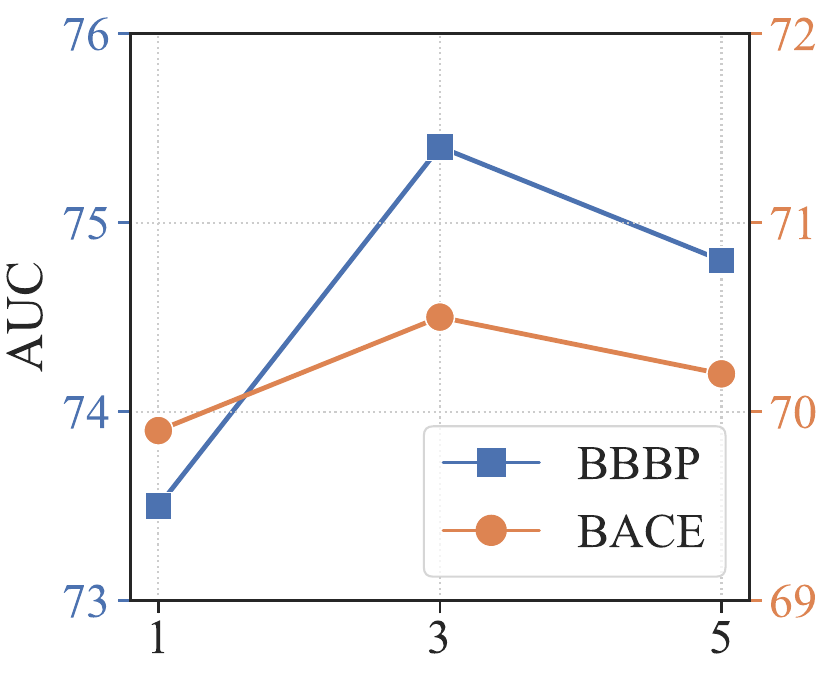}
		\label{fig:c}
	}
 %\hfill
% 	\subfloat[Ablation]{
% 		\includegraphics[width=0.42\linewidth]{figures/ablation.pdf}
% 		\label{fig:abl}
% 	}
% 	\vspace{-2ex}
	\caption{Effect of different hyperparameters of {\ours} for two graph learning datasets.}
% 	\vspace{-4ex}
\label{fig:ablation_graph}
\end{figure}

We use BBBP and BACE as an example to study the effect of different hyperparameters on graph datasets, shown in Figure \ref{fig:ablation_graph}. We find that  for these two datasets, $k=5$ and $c=3$ lead to the best performance. 

\begin{table*}[]
\centering
\begin{tabular}{ccccc}
\toprule 
\bf Hyper-parameter &\bf BBBP& \bf  BACE & \bf Esol & \bf Lipo  \\ \midrule 
Dropout Ratio & \multicolumn{4}{@{\hskip1pt}c@{\hskip1pt}}{0.1}  \\ 
% Maximum Tokens  & 128 & 256 & 256 \\ \hline 
Batch Size for Labeled Data & 16 & 16 & 8 & 8 \\ 
Batch Size for Unlabeled Data  & \multicolumn{4}{@{\hskip1pt}c@{\hskip1pt}}{16} \\ 

Weight Decay & \multicolumn{4}{@{\hskip1pt}c@{\hskip1pt}}{1e-4}  \\
Learning Rate &\multicolumn{4}{@{\hskip1pt}c@{\hskip1pt}}{5e-5}   \\
Initialization Epochs & 10 & 10 & 12 & 12   \\ 
% $T$ &  4000 & 3000 & 4000  \\ \hline
$k$ & 5  & 5  & 3  &  5 \\
$c$ & 3 &  3 &  1 &  3   \\ 
\bottomrule
\end{tabular}
\caption{Hyper-parameter configurations for semi-supervised graph learning.}
\label{tab:hyperparameter-semi-graph}
\end{table*}

\section{Error of Pseudo Labels and Running Time on Graph Datasets}
\label{apd:molecule}
\begin{figure}[t]
    \centering
     \vspace{-1ex}
    \includegraphics[width=0.93\linewidth]{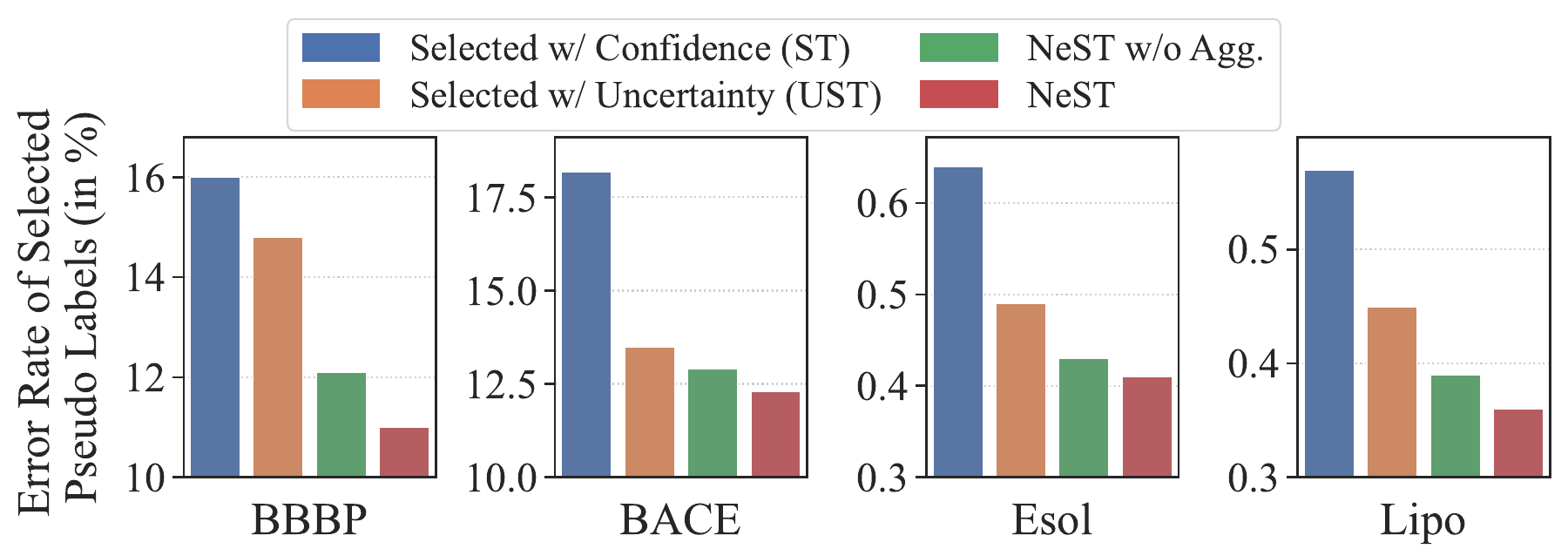}
    \caption{Error rates of pseudo labels selected by different methods. Agg. is the aggregation technique in Section~\ref{sec:selftrain}.}
    \label{fig:case-mol}
\end{figure}

Figure \ref{fig:case-mol} shows the error of pseudo labels for four graph learning datasets. Note that for BBBP and BACE, we show the error rate of pseudo labels, same as the text datasets. 
For Esol and Lipo, since they are regression datasets, we show the average RMSE of the pseudo labels. Note that the lower value indicates higher quality of pseudo labels. The results indicate that our proposed two strategies also improve the quality of pseudo labels.

\begin{figure}[t]
    \centering
    \includegraphics[width=0.93\linewidth]{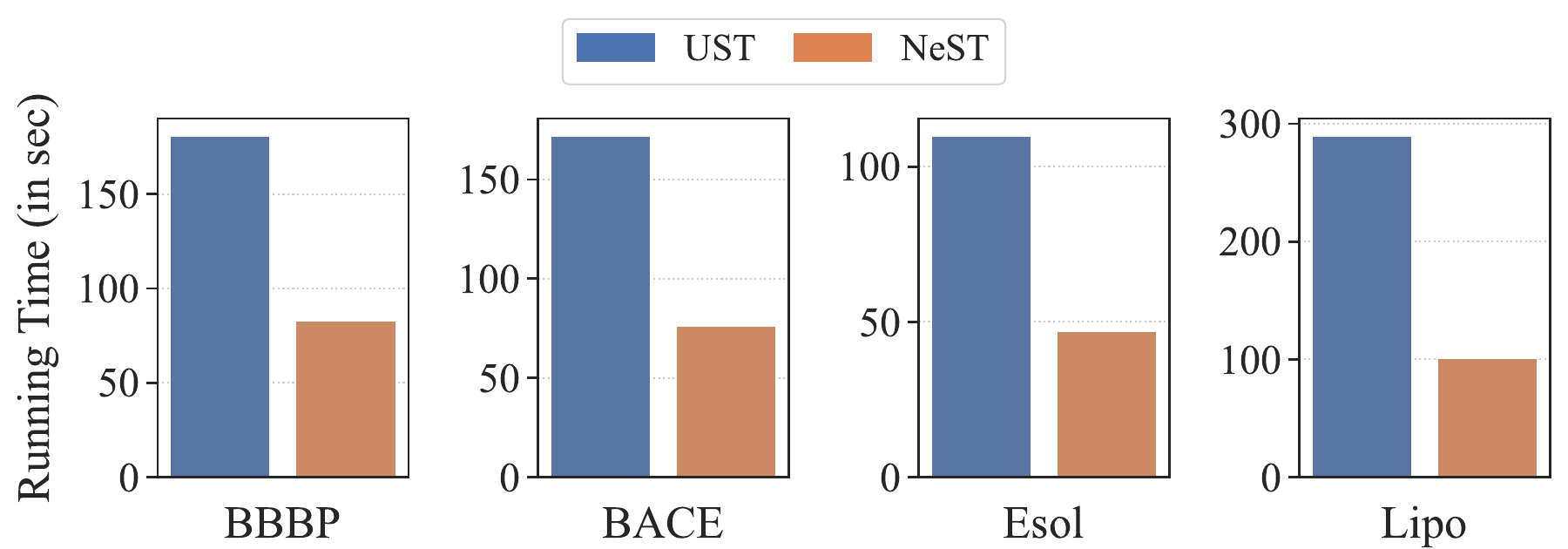}
    \caption{Running time of {\ours} and the best baseline (UST) on four graph datasets.}
    \label{fig:runtime-mol}
    \vspace{-1ex}
\end{figure}

Figure \ref{fig:runtime-mol} shows the running time of the best baseline (UST) and {\ours}. 
Note that CEST is designed for text data, and cannot be directly used for molecular data. 
From the results, we find that {\ours} saves 54.1\%--65.1\% of the running time. We point out that the improvement of running time is rather smaller when compared with the text datasets. 
This is mainly because the size of datasets are much smaller, thus the additional time for multiple inferences in UST is not excessive. 
That being said, these results can well illustrate that {\ours} is more efficient and will serve as a complement to existing self-training approach. 

% \section{Reproducibility}
% The sample code and scripts for running our experiments are submitted to the \emph{supplementary material}. The full code will be release upon the acceptance of this paper. 
\end{document}